%% file: main.tex
\documentclass{article}

\PassOptionsToPackage{table}{xcolor}
\usepackage{iclr2025_conference,times}

\input{math_commands.tex}

\usepackage[T1]{fontenc}
\usepackage[utf8]{inputenc}
\usepackage{latexsym}
\usepackage{marvosym}

\usepackage{amsmath}
\usepackage{amssymb}
\usepackage{amsfonts}
\usepackage{booktabs}
\usepackage{makecell}
\usepackage{multirow}
\usepackage{array}
\usepackage{colortbl}
\usepackage{graphicx}
\usepackage{xspace}
\usepackage{placeins}
\usepackage{caption}
\usepackage[most]{tcolorbox}
\usepackage{enumitem}
\tcbuselibrary{breakable}

\newcommand{\ours}{CuRe\xspace}

\newcommand{\authormark}[1]{\ensuremath{^{#1}}}

\newcommand{\hall}[1]{\textcolor{red}{#1}}
\newcommand{\cureextra}[1]{\textcolor{blue}{#1}}
\newtcolorbox{captioncasebox}[2][]{%
  enhanced,
  colback=white,
  colframe=black!18,
  boxrule=0.3pt,
  arc=0.5mm,
  left=4pt, right=4pt, top=3pt, bottom=3pt,
  width=\linewidth,
  fontupper=\footnotesize,
  before upper={%
    \textbf{\footnotesize #2}\par\vspace{1pt}%
    \noindent\textcolor{black!22}{\rule{\linewidth}{0.25pt}}\par\vspace{2pt}},
  #1
}

\title{Claim-Level Rubric Rewards for Video Caption Reinforcement Learning}

\author{
{\large
Mingqi~Gao\authormark{1,3,*},
Hongyuan~Dong\authormark{3,*,\dagger},
Yifei~Chen\authormark{3,*}, Zhisheng~Zhong\authormark{3},\\
Zheng~Ruan\authormark{3}, Wenjin~Hou\authormark{3},
Yu~Chen\authormark{2}, Han~Hu\authormark{3,\ddagger},
Yansong~Tang\authormark{1,\text{\Letter}}
}\\[1mm]
{\large
\authormark{1}\textbf{Tsinghua Shenzhen International Graduate School, Tsinghua University}
\quad
\authormark{2}\textbf{University of Chinese Academy of Sciences}
}\\
{\large
\authormark{3}\textbf{LLM Department, Tencent}
}\\[1mm]
{\normalsize
\texttt{minkkigao@gmail.com}\quad
\texttt{tang.yansong@sz.tsinghua.edu.cn}}
}

\begin{document}
\maketitle
\begingroup
\renewcommand{\thefootnote}{}
\footnotetext{\hspace{1pt}%
\textsuperscript{*}Equal contribution.\quad
\textsuperscript{\ensuremath{\dagger}}Project leader.\quad
\textsuperscript{\ensuremath{\ddagger}}Project supervisor.\quad
\textsuperscript{\Letter}Corresponding author.}
\endgroup

\input{sections/00_abstract}
\input{sections/01_introduction}
\input{sections/02_related_work}

\input{sections/03_method}

\input{sections/04_experiments}
\input{sections/05_conclusion}
\input{sections/06_discussion_limitations}

\FloatBarrier
\bibliographystyle{iclr2025_conference}
\bibliography{bibtex_candidates}

\input{sections/appendix}

\end{document}

%% file: math_commands.tex
\usepackage{amsmath,amsfonts,bm}
\usepackage{multicol}
\def\eqref#1{equation~\ref{#1}}
\def\1{\bm{1}}

\DeclareMathAlphabet{\mathsfit}{\encodingdefault}{\sfdefault}{m}{sl}
\SetMathAlphabet{\mathsfit}{bold}{\encodingdefault}{\sfdefault}{bx}{n}

%% file: sections/00_abstract.tex
\begin{abstract}
In this paper, we introduce Claim-Level Rubric Rewards (\ours), a structured reward framework designed to address the reward-design bottleneck in reinforcement learning for dense video captioning. Existing reward designs generally fall into two categories: holistic response-level judgment across heterogeneous criteria, or alignment-based evaluation against reference captions. However, both paradigms suffer from fundamental limitations. Holistic rewards struggle to ensure factual accuracy and are prone to stylistic reward hacking, while reference-based rewards overly rely on rigid textual alignment, failing to preserve the completeness and diversity inherent to open-ended generation tasks.
To address these challenges, \ours reformulates reward modeling as fine-grained claim-level verification. Specifically, \ours decomposes captions into category-aware atomic claims through a structured rubric, converting holistic evaluation into simpler and more reliable claim-level verification. 
% To reward captions that are both faithful and informative, we further introduce a reference-anchored calibration mechanism that matches video-supported candidate claims against reference claims. This design suppresses unverified verbosity while still assigning bounded credit to video-supported details beyond a single reference caption, inducing teacher-anchored mode seeking at the reward level.
To reward captions that are both faithful and informative, we further introduce a reference-anchored calibration mechanism that leverages reference claims as salience anchors and visual grounding as factual evidence. This encourages reward-level mode seeking toward vital content, while assigning bounded credit to visual grounded details absent from the reference, preserving descriptiveness without drifting into verbosity.
% while maintaining descriptive breadth by rewarding grounded details beyond the reference and discouraging unsupported verbosity.
Integrated into the standard GRPO paradigm, \ours consistently improves both accuracy and completeness. Across captioning, re-captioning, and caption-to-QA tasks, our 30B-A3B model consistently outperforms its same-scale models and surpasses substantially larger baselines on multiple benchmarks.
\end{abstract}

%% file: sections/01_introduction.tex
\section{Introduction}

Video captioning represents a fundamental capability in video understanding. It serves as an essential data source for video-language alignment, a cornerstone for advanced reasoning~\citep{maaz2024video,yang2024vript}, and a critical enabler for downstream tasks ranging from controllable video generation~\citep{xiong2024lvd} to content recommendation.
In pursuit of models capable of generating detailed and faithful descriptions, the prevailing paradigm relies on Supervised Fine-Tuning (SFT) using high-quality corpora distilled from proprietary models or human annotation~\citep{Yuan2025Tarsier2,Chen2024ShareGPT4Video}. Despite its ubiquity, this approach suffers from distinct bottlenecks. Beyond the exorbitant cost of data curation, SFT inherently suffers from exposure bias~\citep{Arora2022exposure,Song2026survey}: the model is trained exclusively on teacher-generated prefixes and scarcely learns to recover from its own deviations at inference, with errors compounding over dense captions. Moreover, the paradigm tends to capture the teacher's style and confidence rather than its factual grounding~\citep{Gudibande2023imitating}.

\input{figures/figure1_teaser}

Reinforcement Learning (RL) offers a viable alternative by optimizing the captioner against reward signals derived from its own rollouts. Current approaches design rewards along two axes: \textbf{holistic scoring}, which prompts a reward model to directly evaluate the generated text against coarse criteria~\citep{Ye2025JudgeBias,Xing2026CapRL}, and \textbf{reference-grounded matching}, which aligns the rollout against a ground-truth reference via lexical metrics (e.g., ROUGE, CIDEr)~\citep{Pasunuru2017EntailmentRewards,Oliveira2021CIDErR} or model judgments~\citep{Meng2025VideoCapR1,Zhong2025OwlCap}. While these methods report consistent gains over their SFT counterparts, two fundamental challenges remain largely unaddressed: 
\textbf{(i) Lack of Rubric Guidance:} Holistic scoring collapses multifaceted visual dimensions (e.g., motion, interaction) into a single scalar. This coarse-grained feedback strains the reward model's reasoning capacity and invites reward hacking, where the model exploits stylistic preferences rather than factual grounding.
\textbf{(ii) Rigid Reference Alignment:} Given the open-ended nature of video description, anchoring rewards strictly to reference text cannot faithfully reflect factual correctness. Valid descriptions that capture the video's semantics but deviate from the reference's specific phrasing are often heavily penalized.

To bridge these gaps, we propose \textbf{C}laim-Level R\textbf{u}bric \textbf{Re}wards (\textbf{\ours}), a comprehensive reward framework that redefines fine-grained evaluation feedback for video captioning.
First, \ours operationalizes a fine-grained Rubric that systematically decomposes captions across diverse visual categories and resolves them into atomic claims. By projecting captions into this rubric matrix, \ours converts an otherwise intractable holistic judgment into simpler, visual atomic verifications, significantly lowering the task complexity for the validator and ensuring comprehensive feedback coverage.
Second, to achieve robust rollout evaluation, we couple the aforementioned visual verification with one-time matching to reference claims through a \emph{reference-anchored calibration mechanism}.
Instead of forcing rigid alignment, the reference claims anchor full credit for matched claims, while unmatched but video-supported residual claims receive only attenuated credit, thereby diluting the relative incentive for non-essential or verbose descriptions.  
This mechanism induces reference-anchored mode seeking at the reward layer, encouraging the model to discover faithful, informative descriptive patterns while circumventing reward hacking.

We validate \ours on public captioning and video hallucination benchmarks, demonstrating that our reward-trained Qwen3-VL captioner significantly improves both descriptive density and factual faithfulness over its SFT counterpart. Furthermore, re-captioning three pretraining datasets with \ours consistently boosts downstream performance across most metrics. Finally, a Prism-style caption-to-QA evaluation~\citep{qiao2024prism} confirms that the \ours-trained captioner outperforms substantially larger open-source baselines and rivals the advanced proprietary models.

Our key contributions can be summarized as follows:
\begin{itemize}[leftmargin=1.5em]

\item We propose \textbf{\ours}, a structured reward design that decomposes video captions into category-specific, atomic claims. By evaluating these fine-grained claims individually, \ours provides precise feedback and ensures comprehensive reward coverage.
  
\item We design a reference-anchored calibration mechanism that matches supported rollout claims once against reference claims and assigns attenuated credit to video-supported residuals, effectively encouraging the model to generate faithful, informative descriptions instead of verbose trivial patterns.
  
\item Extensive experiments demonstrate that \ours simultaneously improves factual faithfulness and descriptive density, outperforming much larger open-source models and boosting downstream training performance.
\end{itemize}

%% file: figures/figure1_teaser.tex
\begin{figure}[t]
    \centering
    \includegraphics[width=\linewidth]{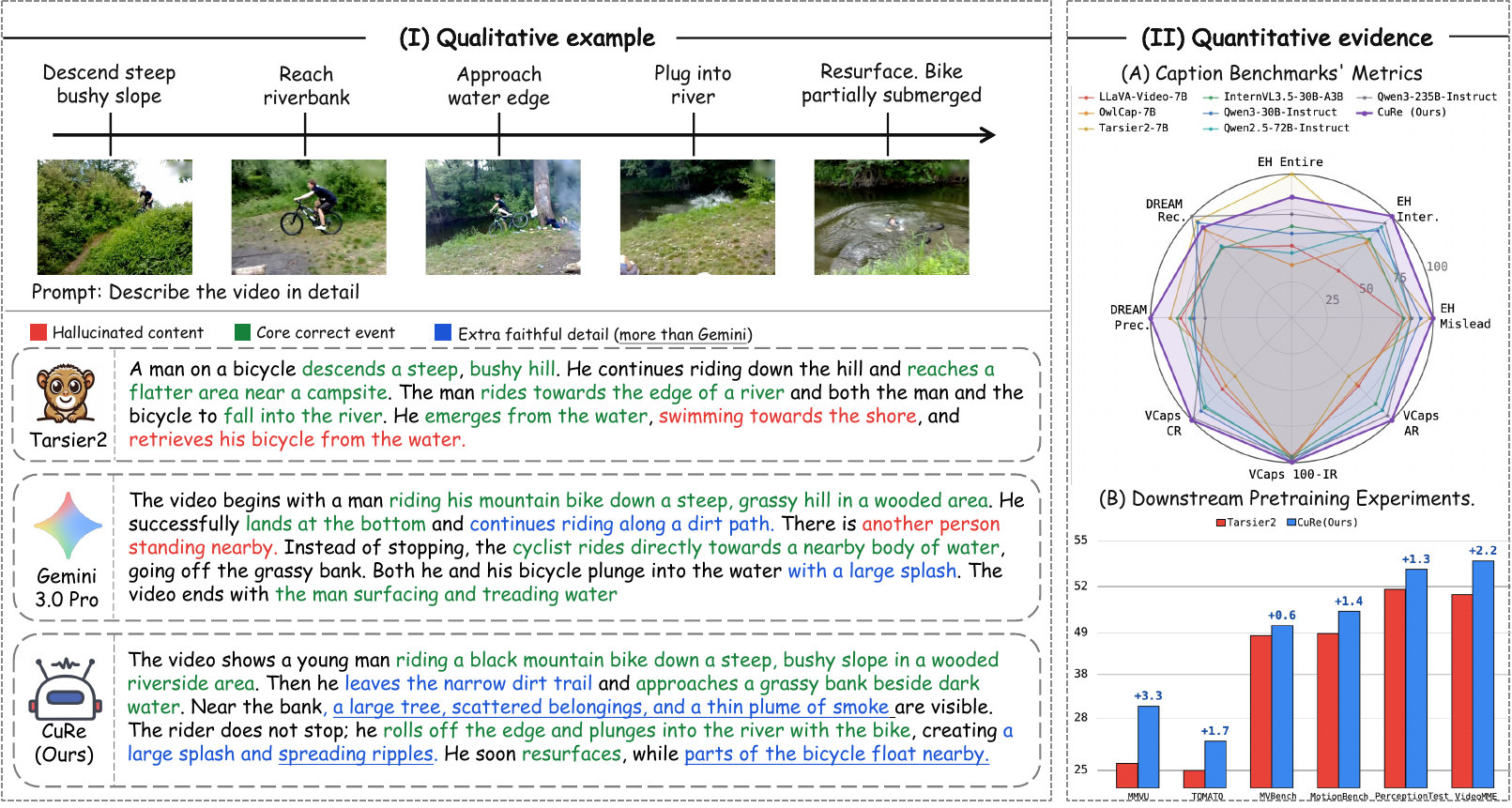}
    \caption{Qualitative and quantitative results of \ours. \ours produces faithful and detailed captions by retaining core events, reducing hallucinations, and adding visually supported details, as shown by qualitative examples in (I) and quantitative comparisons in (II).}
    \label{fig:teaser}
    \vspace{-2mm}
\end{figure}

%% file: sections/02_related_work.tex
\section{Related Work}
\subsection{Video Captioning}
Video captioning has shifted from producing concise summaries to generating detailed, multi-axis descriptions. Early paradigms predominantly rely on supervised fine-tuning, leveraging either human-annotated datasets~\citep{xu2016msr,chen2011msvd,wang2019vatex} or distilled corpora from proprietary models~\citep{Chen2024ShareGPT4Video,Zhang2024LLaVAVideo}. To further scale up supervision, recent methods employ fine-tuned captioners to self-generate massive training data to explore the model capacity limits~\citep{Wang2024Tarsier,Yuan2025Tarsier2,cho2026perceptionlm}. 
More recently, reinforcement learning (RL) has emerged as a promising alternative. Several pioneering works~\citep{Ahn2024VLMRLAIF,tang2025avc,Meng2025VideoCapR1} have integrated advanced RL paradigms (e.g., PPO, DPO, and GRPO) into video captioning task.
However, these methods depend on coarse-grained, caption-level feedback, such as global preferences or holistic model evaluations. Such feedback requires substantial reasoning from the judge model and can induce reward hacking.

\subsection{Rubric Rewards for Reinforcement Learning}
Rubric-based rewards have recently emerged as an effective paradigm for reinforcement learning on open-ended generation tasks, where holistic scalar rewards often fail to capture diverse quality dimensions. Instead of relying on single-score judgments, rubric rewards decompose evaluation into explicit and independently assessable criteria, providing more structured supervision signals for policy optimization.
Recent works explore rubric generation and scalable reward construction for open-ended RL, including Chain-of-Rubrics in RM-R1~\citep{Chen2025RMR1}, dynamically generated checklist rewards in RaR~\citep{Gunjal2025RaR}, and scalable rubric synthesis frameworks such as OpenRubrics, Auto-Rubric, and RLCER~\citep{Liu2025OpenRubrics,Xie2025AutoRubric,Sheng2026RLCER}. More recently, rubric-guided RL has been extended to multimodal generation, where RubiCap applies rubric-based rewards to dense image captioning~\citep{Huang2026RubiCap}. Our work further extends rubric rewards to dense video captioning by introducing category-typed atomic visual claims as fine-grained scoring units, enabling more precise visual credit assignment for open-ended video description.

%% file: sections/03_method.tex
\section{Methodology}
\label{sec:method}

% To relieve holistic caption-level scoring from conflating length, fluency, and correctness, we realize the claim-level rubric reward of \ours as a three-stage pipeline, which is shown in  Figure~\ref{fig:method_overview}. We project a rollout into typed atomic visual claims, verify each claim against the video, and match supported claims once to reference claims of the same category (\S\ref{subsec:construction}). A teacher caption rarely enumerates all caption-worthy content. Treating the teacher caption as a partial salience anchor rather than an exhaustive target, we assign selective rubric credit: full credit to teacher-matched claims and bounded credit to unmatched but video-supported residual claims (\S\ref{subsec:scoring}). Precision and recall are aggregated into a scalar reward (\S\ref{subsec:aggregation}) consumed by standard Group-Relative Policy Optimization.%without modification.
Our goal is to build a rubric reward system for video captioning that provides more precise and interpretable feedback than holistic caption-level scoring, encouraging captions that are both factually faithful and descriptively dense. To this end, we propose \ours, a three-stage claim-level reward pipeline, as illustrated in Figure~\ref{fig:method_overview}. First, we decompose rollout captions and reference captions into category-specific atomic visual claims, verify candidate claims against the video, and establish matches between candidate claims and reference ones (\S\ref{subsec:construction}). Second, we introduce an adaptive reward calibration stage that balances visual factual grounding with reference-claim alignment when assigning credit (\S\ref{subsec:scoring}). Finally, we aggregate the calibrated category-wise scores into a single scalar and combine it with a length-control penalty to form the final reward (\S\ref{subsec:aggregation}).

\subsection{Claim-Level Rubric Construction}
\label{subsec:construction}

CuRe first decomposes each caption into atomic, verifiable visual claims. Given a video $x$ and a model-rollout caption $y$, an LLM-based decomposer~\citep{QwenTeam2025Qwen3VL} projects the caption into a set of typed candidate claims:
\begin{equation}
\mathcal{C} = \operatorname{Decompose}(y) = \{(c_i, k_i)\}_{i=1}^{n},
\label{eq:claim-decomposition}
\end{equation}
where $c_i$ is an atomic visual claim, $k_i \in \mathcal{K}$ denotes its visual category under the schema illustrated in Appendix~\ref{app:priors}, and $n$ is the number of extracted claims. We also decompose a reference caption $y^\star$ generated by an advanced proprietary VLM~\citep{google2025gemini}, yielding the reference claim set $\mathcal{C}^{\star} = \operatorname{Decompose}(y^\star)$.

CuRe then verifies and matches claims within each visual category. For a category $k$, let $\mathcal{C}_k$ and $\mathcal{C}_k^\star$ denote the candidate and reference claims assigned to $k$, respectively. A VLM~\citep{QwenTeam2025Qwen3VL} video verifier checks each candidate claim in $\mathcal{C}_k$ against the video and retains the supported subset:
\begin{equation}
\mathcal{S}_k =
\{c \in \mathcal{C}_k \mid v(x,c)=1\},
\label{eq:supported-subset}
\end{equation}
where $v(x,c)$ is the VLM verifier indicating whether claim $c$ is supported by video $x$. The supported claims are then matched to reference claims within the same category:
\begin{equation}
\mathcal{M}_k =
\{(c, c^\star) \in \mathcal{S}_k \times \mathcal{C}_k^\star
\mid m(c,c^\star)=1\},
\label{eq:teacher-matching}
\end{equation}
where $m(c,c^\star)$ indicates that the two claims are semantically aligned, and each claim can appear in at most one matched pair.
We introduce basic claim-level precision and recall scores:
\begin{equation}
p_k =
\frac{|\mathcal{S}_k|}{|\mathcal{C}_k|},
\qquad
r_k =
\frac{|\mathcal{M}_k|}{|\mathcal{C}_k^\star|}.
\label{eq:base-pr}
\end{equation}
Here, precision measures factual grounding of the generated caption, while recall measures how much reference content is recovered by grounded candidate claims.

\input{figures/figure2_method_overview}

\subsection{Reference-Anchored Calibration}
\label{subsec:scoring}
The base claim-level scores in Eq.~\ref{eq:base-pr} provide a simple starting point for measuring factual grounding and reference coverage. However, reference claims provide asymmetric evidence: they are reliable anchors for salient, high-confidence content, but should be treated as a partial view rather than a complete account of the video's grounded evidence. This motivates \emph{reference-anchored calibration}, which assigns preferential credit to reference-matched claims while allowing verified residual claims to contribute in a controlled manner.

For precision, CuRe replaces uniformly support-based credit with reference-aware credit. Reference-matched supported claims receive full credit, while unmatched supported claims receive attenuated credit $w_u \in [0,1]$:
\begin{equation}
p_k^{\mathrm{CuRe}} =
\frac{
|\mathcal{M}_k| + w_u\left(|\mathcal{S}_k|-|\mathcal{M}_k|\right)
}
{
|\mathcal{C}_k|
}.
\label{eq:cure-precision}
\end{equation}
This favors claims aligned with salient reference content, while retaining discounted credit for additional details grounded in the video.

For recall, CuRe avoids treating the reference claims as a closed coverage target. Unmatched supported claims may expand the effective target, but only through bounded attenuated credit:
\begin{equation}
r_k^{\mathrm{CuRe}} =
\frac{
|\mathcal{M}_k| +
w_u \min\left(|\mathcal{S}_k|-|\mathcal{M}_k|,\; \delta_k\right)
}
{
|\mathcal{C}_k^\star| +
w_u \min\left(|\mathcal{S}_k|-|\mathcal{M}_k|,\; \delta_k\right)
},
\label{eq:cure-recall}
\end{equation}
where $\delta_k = |\mathcal{C}_k^\star|-|\mathcal{M}_k|$ is the coverage bound. The cap operation restricts residual credit by this bound, preventing recall-side reward hacking through verbose unmatched claims while still allowing faithful evidence beyond the reference to contribute.
Without this cap operation, our experiments show that the video captioner may exploit the recall metric by generating correct yet redundant visual details that are not covered by the reference caption.

\subsection{Final Reward Aggregation}
\label{subsec:aggregation}
After obtaining calibrated per-category scores, CuRe aggregates precision and recall over their respective active categories. Specifically, precision and recall are computed over non-empty candidate categories $\mathcal{K}_{\mathrm{cand}} = \{k \in \mathcal{K} : |\mathcal{C}_{k}| > 0\}$ and non-empty reference categories $\mathcal{K}_{\mathrm{ref}} = \{k \in \mathcal{K} : |\mathcal{C}_{k}^{\star}| > 0\}$, respectively. By weighting with category priors $\omega_k$, the aggregated metrics are defined as:
\begin{equation}
p^{\mathrm{CuRe}} = \frac{\sum_{k \in \mathcal{K}_{\mathrm{cand}}} \omega_k p_{k}^{\mathrm{CuRe}}}{\sum_{k \in \mathcal{K}_{\mathrm{cand}}} \omega_k}, 
\qquad
r^{\mathrm{CuRe}} = \frac{\sum_{k \in \mathcal{K}_{\mathrm{ref}}} \omega_k r_{k}^{\mathrm{CuRe}}}{\sum_{k \in \mathcal{K}_{\mathrm{ref}}} \omega_k}.
\label{eq:aggregated-pr}
\end{equation}
Notably, $p^{\mathrm{CuRe}}$ is set to $0$ when $\mathcal{K}_{\mathrm{cand}}=\varnothing$, ensuring that empty captions do not receive vacuous precision credit.
We define the overall CuRe reward as a weighted combination of the aggregated precision and recall:
\begin{equation}
R^{\mathrm{CuRe}} = \lambda\,p^{\mathrm{CuRe}} + (1 - \lambda)\, r^{\mathrm{CuRe}},
\label{eq:cure-reward}
\end{equation}
where $\lambda \in [0,1]$ is a hyperparameter that controls preference between conservative and informative captioning tendencies. Moreover, to discourage over-generation, we incorporate a length penalty $\Omega$ into the final reward, with details provided in Appendix~\ref{app:training_config}.

For training, we adopt the standard GRPO~\citep{shao2024deepseekmath} framework with CuRe as the reward function. For each video, the policy model samples a group of eight captions, which are scored by evaluating its candidate claims against the video and reference claims, together with the length-control penalty. This claim-level reward provides fine-grained and coverage-aware feedback, steering optimization toward captions that are factually faithful and descriptively dense, rather than verbose but trivial.

%% file: figures/figure2_method_overview.tex
\begin{figure*}[t]
\centering
\includegraphics[width=\linewidth]{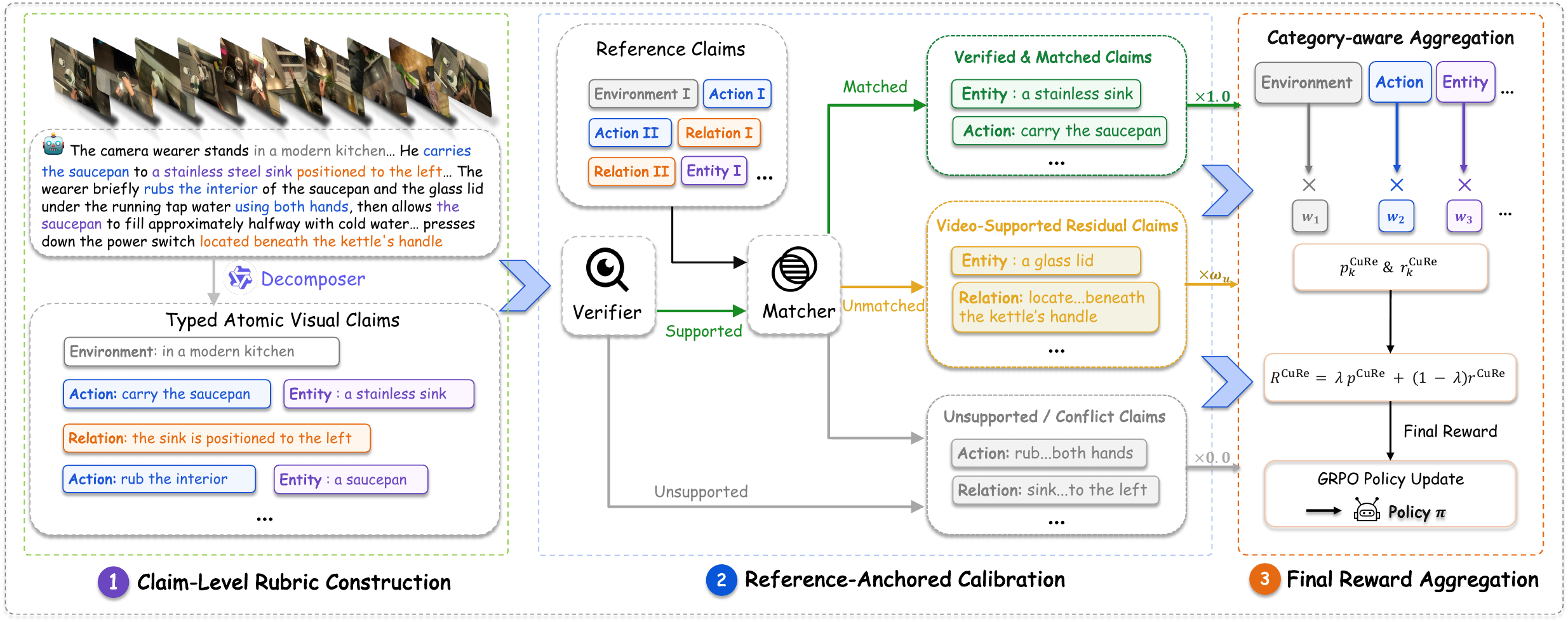}
% \caption{Overview of the \ours framework.
% \ours decomposes dense captions into atomic visual claims, verifies them against the video, and matches supported claims once against reference claims. A selective rubric then assigns full credit to reference-matched claims, attenuated credit to unmatched but video-supported residual claims, and zero credit to unsupported claims. }
% \label{fig:method_overview}
\caption{Overview of the \ours framework.
\ours first decomposes dense captions into category-specific atomic visual claims, verifies candidate claims against the video, and matches video-supported claims once against reference ones. It then applies reference-anchored reward calibration, assigning higher credit to reference-aligned supported claims, bounded credit to supported details absent from the reference, and zero credit to unsupported or conflicted claims. The calibrated category-wise scores are finally aggregated into the overall reward.}
\label{fig:method_overview}
\end{figure*}

%% file: sections/04_experiments.tex
\section{Experiments}
\input{tables/table1_hallusion_caption_bmks}
% In this section, we show the effectiveness of \ours through caption quality evaluation (\S\ref{subsec:open-source-caption-benchmarks}), VLM pretraining validation (\S\ref{subsec:downstream-pretraining}), and general benchmark evaluation with caption proxies (\S\ref{subsec:prism-caption-to-qa}). 
% Section~\ref{subsec:ablation} further validates the effectiveness of our reward design by ablating individual components, and \S\ref{subsec:qualitative-analysis} gives a qualitative example.

In this section, we evaluate \ours from three perspectives: direct caption quality on open-ended captioning benchmarks (\S\ref{subsec:open-source-caption-benchmarks}),  downstream task performance after \ours-enhanced pretraining (\S\ref{subsec:downstream-pretraining}),  and caption-to-QA evaluation across multiple benchmarks (\S\ref{subsec:prism-caption-to-qa}). We further analyze the contribution of each reward component through ablations (\S\ref{subsec:ablation}) and provide qualitative examples in \S\ref{subsec:qualitative-analysis}.

% We organize the experiments around two questions. First, does \ours improve dense video caption quality and reduce event-level hallucination, and which reward components drive the gains? We answer this by scoring our captions on public caption-quality and hallucination benchmarks (\S\ref{subsec:open-source-caption-benchmarks}), and by ablating individual reward components (\S\ref{subsec:ablation}). Second, are the resulting captions useful for downstream applications? We answer this in two settings: \emph{downstream pretraining utility} (\S\ref{subsec:downstream-pretraining}) and \emph{Prism-style caption-to-QA utility} (\S\ref{subsec:prism-caption-to-qa}).

\subsection{Experimental Setup}
\label{subsec:setup}

\paragraph{Implementation Details.} We initialize the policy captioner from Qwen3-VL-30B-A3B-Instruct~\citep{QwenTeam2025Qwen3VL} and post-train it on a Gemini-3.0-Pro-distilled caption corpus drawn from open-source web videos~\citep{Yuan2025Tarsier2}. The captioner is first trained with a 78{,}144-sample video caption dataset as SFT warmstart, and then optimized on a deduplicated 8{,}000-video subset with GRPO. In the reward runtime, Qwen3-VL-235B-A22B-Instruct decomposes only the rollout captions, verifies rollout claims against the video, and performs the one-to-one matcher under task-specific prompts (Appendix~\ref{app:prompts}). Gemini-3.0-Pro~\citep{google2025gemini} provides the reference captions and their decomposed reference claims $\mathcal{C}^{\star}$. We then perform one semantic match between supported rollout claims and these reference claims. We set $w_u = 0.5$ and $\lambda = 0.7$, and refer to Appendix~\ref{app:training_config} for detailed training settings. None of the SFT or GRPO videos overlap with our evaluation benchmarks.

\paragraph{Baselines.} To show the leading performance of \ours, we compare our captioner not only with open-source video captioners such as Tarsier~\citep{Wang2024Tarsier,Yuan2025Tarsier2}, OwlCap~\citep{Zhong2025OwlCap}, and LLaVA-Video, but also with general-purpose VLMs including InternVL3.5, Qwen2.5-VL, and Qwen3-VL~\citep{QwenTeam2025Qwen3VL}.
We also include proprietary models, including Seed 2.0 Pro and Gemini 3.0 Pro, as reference baselines to assess how far \ours narrows the gap between open-source video captioners and leading proprietary VLMs.
% To demonstrate the leading performance of \ours, we select not only open-source video captioners and VLMs, including Tarsier family~\citep{Wang2024Tarsier,Yuan2025Tarsier2}, OwlCap~\citep{Zhong2025OwlCap}, LLaVA-Video, InternVL3.5, Qwen2.5-VL, and Qwen3-VL~\citep{QwenTeam2025Qwen3VL}, but also proprietary models such as Seed 2.0 Pro and Gemini 3.0 Pro as baselines. 

% \paragraph{Evaluation.} 

% \paragraph{Evaluation and Baselines.} Our baselines come from three families. 1. \emph{Open-source video captioners specialized for dense or motion-detail captioning} include the Tarsier family~\citep{Wang2024Tarsier,Yuan2025Tarsier2}, OwlCap~\citep{Zhong2025OwlCap}, and LLaVA-Video. 2. \emph{General-purpose open-source VLMs} include the InternVL family, Qwen2.5-VL, and the Qwen3-VL family~\citep{QwenTeam2025Qwen3VL}. 3. \emph{Proprietary models} (Seed 2.0 Pro and Gemini 3.0 Pro) are included only as reference systems and are excluded from open-source highlighting. We cite originally reported benchmark results in \ref{tab:video_hallusion_compact_results} when available and otherwise re-evaluate available models with the official evaluator. %Full prompts and reference sources are listed in Appendix~\ref{app:baselines}.

\subsection{Caption Quality Evaluation}
\label{subsec:open-source-caption-benchmarks}
In this part, we evaluate \ours on public video captioning benchmarks to assess the quality of the generated captions.

\paragraph{Settings.} We evaluate caption quality on three benchmarks. EventHallusion~\citep{Zhang2024EventHallusion} probes event hallucination under Entire, Interleave, and Misleading settings, where GPT-based description matching judges whether generated captions are consistent with the target event. We report its description-level scores. VCapsBench~\citep{Zhang2025VCapsBench} evaluates captions through text-only QA over human-verified video questions with ternary answers, reporting Accuracy Rate (AR), Inconsistency Rate (IR, lower is better), and Coverage Rate (CR). DREAM-1K \citep{Wang2024Tarsier} uses AutoDQ to extract events from generated and reference descriptions and compute event-level precision, recall, and F1.

\paragraph{Results and analysis.} 
% Among open-source models, \ours gives the strongest overall direct-captioning performance in Table~\ref{tab:video_hallusion_compact_results}. On EventHallusion, the Overall description-level score reaches 69.5, 10.6 ahead of its same-sized Qwen3-VL-30B-A3B-Instruct base and 6.2 ahead of Tarsier2-7B, the previous open-source best. 
% The pattern is robust across baseline types: \ours beats both the much larger Qwen3-VL-235B-A22B-Instruct and even the proprietary Seed 2.0 Pro on all three VCapsBench axes, and beats other open-source models on EventHallusion Overall. The gains span factual faithfulness (EventHallusion Overall, VCapsBench IR, etc.) and descriptive density (VCapsBench CR, DREAM-1K Recall). On DREAM-1K, \ours leads on precision and F1, with recall 4.5 points behind Qwen3-VL-235B-A22B-Instruct, which indicates a trade-off favoring faithful, informative descriptions over recall padding. 
Among open-source models, \ours achieves the strongest overall direct-captioning performance as shown in Table~\ref{tab:video_hallusion_compact_results}. On EventHallusion, \ours obtains the best overall score of 69.5, improving over its same-sized Qwen3-VL-30B-A3B-Instruct baseline by 10.6 points and surpassing the previous open-source best, Tarsier2-7B, by 6.2 points. 
On VCapsBench, \ours ranks first among open-source models on all three axes, indicating that its captions are both more answerable and less inconsistent. It also outperforms the much larger Qwen3-VL-235B-A22B-Instruct and the proprietary Seed 2.0 Pro across all VCapsBench metrics. 
On DREAM-1K, \ours achieves the best precision and F1 among open-source models, while its recall is 4.5 points lower than Qwen3-VL-235B-A22B-Instruct. 
Overall, \ours demonstrates remarkable performance gain on factual faithfulness (EventHallusion Overall, VCapsBench IR, DREAM-1K Prec.), while the descriptive density (VCapsBench CR, DREAM-1K Rec.) is competitive with leading baseline models. 
This suggests that \ours favors faithful and discriminative event descriptions rather than increasing recall through potentially noisier detail coverage.

% \subsection{CuRe-Annotated Caption for VLM Pretraining}
\subsection{Downstream Evaluation via \ours Pretraining}
\label{subsec:downstream-pretraining}
\input{tables/table2_downstream_training}
In this part, we apply model-generated video captions in end-to-end VLM pretraining and show the final benchmark score as a proxy of caption quality evaluation. 
% We further evaluate caption quality through downstream pretraining utility, where the model-generated video captions are used for end-to-end VLM pretraining.
% and treat the resulting benchmark performance as an indirect measure of caption data quality, following the utility-oriented evaluation perspective in prior caption-data studies~\citep{Chen2024ShareGPT4V,Xing2026CapRL}. 
This setup highlights the practical value of \ours for modern VLM data engineering.
\paragraph{Settings.} 
% Beyond scoring well as captions, do \ours-generated captions also serve as stronger supervision when used to pretrain video-language models? To address this problem, 
% We start with Qwen3 4B, Qwen3 ViT, and a randomly initialized projector, followed by a three-stage training protocol from ShareGPT4V~\citep{Chen2024ShareGPT4V,Chen2024ShareGPT4Video} and CapRL~\citep{Xing2026CapRL}. Stage 1 aligns on BLIP-3-Kale~\citep{Awadalla2024BLIP3Kale}, Stage 2 further pretrains on each candidate caption corpus, and Stage 3 performs SFT on a fixed VideoQA mixture of Molmo2-AMA~\citep{Clark2026Molmo2} and LLaVA-Video~\citep{Zhang2024LLaVAVideo}. Each paired comparison changes only the Stage-2 caption source, replacing the original captions with \ours-recaptioned versions on three Stage-2 corpora: Molmo2-Cap, Tarsier2-Recap~\citep{Yuan2025Tarsier2}, and LLaVA-Video-Caption. These three corpora are disjoint from the Stage 3 mixture. Training details are provided in Appendix~\ref{ app:downstream-training}.
We initialize the model with a Qwen3-4B~\citep{yang2025qwen3} language backbone, the Qwen3 vision encoder~\citep{QwenTeam2025Qwen3VL}, and a randomly initialized projector, and train it with a three-stage protocol following ShareGPT4V~\citep{Chen2024ShareGPT4V,Chen2024ShareGPT4Video} and CapRL~\citep{Xing2026CapRL}. Stage-1 performs image-text alignment on BLIP-3-KALE~\citep{Awadalla2024BLIP3Kale}; Stage-2 continues caption pretraining on a candidate caption corpus; and stage-3 applies SFT on a fixed VideoQA mixture from Molmo2-AMA~\citep{Clark2026Molmo2} and LLaVA-Video~\citep{Zhang2024LLaVAVideo}. 
% Each paired comparison changes only the Stage-2 caption source, replacing the original captions with \ours-recaptioned versions on three Stage-2 corpora: Molmo2-Cap, Tarsier2-Recap~\citep{Yuan2025Tarsier2}, and LLaVA-Video-Caption. These three corpora are disjoint from the Stage 3 mixture. Training details are provided in Appendix~\ref{app:downstream-training}.
We reannotate Molmo2-Cap, Tarsier2-Recap~\citep{Yuan2025Tarsier2}, and LLaVA-Video-Caption with \ours, and compare the final VLM performance trained with the original and reannotated corpora in Stage-2.
% training for fair comparison with open-source baselines. 
Additional training details are provided in Appendix~\ref{app:downstream-training}.

\paragraph{Benchmarks.} 
We evaluate downstream video-language performance on six complementary benchmarks: MMVU for expert-level video understanding and reasoning~\citep{Zhao2024MMVU}, MVBench for temporal perception and cognition~\citep{Li2024MVBench}, MotionBench for fine-grained motion understanding~\citep{Hong2024MotionBench}, TOMATO for visual temporal reasoning~\citep{Shangguan2024TOMATO}, and VideoMME, PerceptionTest for broad real-world video understanding covering diverse domains, video lengths, and perception-to-reasoning question types~\citep{Fu2024VideoMME,Patraucean2023PerceptionTest}.
% We evaluate downstream video-language performance on six complementary benchmarks that cover different aspects of video understanding. MMVU evaluates expert-level video understanding and multi-step reasoning abilities~\citep{Zhao2024MMVU}; MVBench focuses on temporal perception and video-centric cognition~\citep{Li2024MVBench}; MotionBench targets fine-grained motion understanding, where models must distinguish subtle motion patterns and dynamic state changes~\citep{Hong2024MotionBench}; TOMATO stresses visual temporal reasoning, especially event ordering, duration, and temporal relation understanding~\citep{Shangguan2024TOMATO}. We further include VideoMME and PerceptionTest as broad real-world video understanding benchmarks, covering diverse domains, video lengths, and perception-to-reasoning question types~\citep{Fu2024VideoMME,Patraucean2023PerceptionTest}.
% To isolate the effect of caption quality, each paired comparison changes only the Stage-2 caption source by replacing the original captions with \ours-recaptioned versions for Molmo2-Cap, Tarsier2-Recap~\citep{Yuan2025Tarsier2}, and LLaVA-Video-Caption. These Stage-2 corpora are disjoint from the Stage-3 VideoQA mixture. Additional training details are provided in Appendix~\ref{app:downstream-training}.

\paragraph{Results and analysis.} 
% As shown in Table~\ref{tab:downstream_pretraining_utility}, training with the \ours-annotated version in stage-2 improves downstream video-language learning on most benchmarks. Averaging each benchmark's overall performance, substituting the original captions with \ours-generated ones demonstrates 2.18, 1.75, and 2.04 point performance gains on Molmo2-Cap, Tarsier2-Recap, and LLaVA-Video-Caption, respectively. These results show that \ours captions provide stronger video-language pretraining supervision rather than merely matching the style of a particular evaluator.
% As shown in Table~\ref{tab:downstream_pretraining_utility}, replacing the original Stage-2 captions with \ours-generated ones consistently improves downstream video-language learning. Averaged over all evaluation benchmarks, \ours yields gains of 2.18, 1.75, and 2.04 points when applied to Molmo2-Cap, Tarsier2-Recap, and LLaVA-Video-Caption, respectively. 
% Since the Stage-3 VideoQA mixture is fixed across all paired comparisons, these improvements isolate the contribution of caption quality during Stage-2 pretraining. 
% The consistent gains across three distinct caption corpora further suggest that \ours provides generally stronger video-language supervision, rather than merely overfitting to the style or bias of a particular evaluator.

As shown in Table~\ref{tab:downstream_pretraining_utility}, replacing the original Stage-2 captions with \ours-generated captions consistently improves downstream video-language learning across different caption corpora. Averaged over all evaluation benchmarks, \ours brings gains of 2.18, 1.75, and 2.04 points on Molmo2-Cap, Tarsier2-Recap, and LLaVA-Video-Caption, respectively. These improvements are broad rather than concentrated on a single benchmark: \ours outperforms the original captions on 25 out of 27 reported metrics, with the only regressions appearing on MMVU for Molmo2-Cap and MVBench for LLaVA-Video-Caption. 

% Since the Stage-3 VideoQA mixture and the remaining training recipe are fixed within each paired comparison, the observed improvements isolate the contribution of Stage-2 caption quality. The consistent improvement on the three corpora with different source videos, dataset scales, and annotating models further indicates that \ours provides generally stronger video-language supervision, rather than merely overfitting to the style or bias of a particular evaluator.
% Instead, claim-level video-grounded recaptioning appears to enhance the pretraining utility of video caption data in a general way, producing captions that transfer more effectively to downstream video-language tasks.
Since the Stage-3 VideoQA mixture and the remaining training recipe are fixed within each paired comparison, the observed improvements isolate the contribution of Stage-2 caption quality. The consistent gains across three corpora with different source videos, scales, and annotating models further indicate that \ours provides stronger video-language supervision, rather than merely overfitting to a particular evaluator.

\subsection{Caption-to-QA Prism Evaluation}
\label{subsec:prism-caption-to-qa}
\input{tables/table3_prism}

We further adopt the Prism framework~\citep{qiao2024prism} for caption quality evaluation, which judges whether \ours captions retain enough information to serve as a video proxy in various video question answering (VQA) scenarios.

\paragraph{Settings.} 
% We adopt the caption-to-QA protocol of the Prism Framework~\citep{qiao2024prism} to evaluate whether \ours captions retain enough information to serve as a video proxy at inference time. 
% The Prism framework decouples VQA into two stages: in Stage-1, each captioner generates a caption for the video, and in Stage~2, a fixed QA model (Seed 2.0 Pro) answers the benchmark question conditioned only on that caption. Scores come from each benchmark's official evaluator, so the numbers reflect caption informativeness as exploited by a single, fixed answering model.
Prism evaluates caption utility via a two-stage VQA pipeline: each captioner first generates a video caption, and a fixed QA model, Seed 2.0 Pro, answers benchmark questions using only that caption. Official benchmark evaluators compute the final scores, measuring how much task-relevant information each caption exposes to the same answerer. We use the same set of benchmarks as introduced in Section~\ref{subsec:downstream-pretraining} to evaluate the video caption quality on comprehensive VQA tasks. 

\paragraph{Results and analysis.} 
Table~\ref{tab:prism_model_benchmark_results} shows that \ours delivers the strongest open-source caption-to-QA utility under the Prism protocol. Averaged over the six benchmark Overall scores, \ours obtains 53.67, surpassing the same-sized Qwen3-VL-30B-A3B-Instruct by 5.60 points and the much larger Qwen3-VL-235B-A22B-Instruct by 4.83 points. 
% This advantage is not limited to a single dataset: 
\ours ranks first among open-source captioners on 10 out of 11 reported columns, indicating that \ours-generated captions preserve more useful information for varying video QA scenarios.

The improvement is particularly pronounced on long-form video understanding. On VideoMME Long, \ours outperforms Qwen3-VL-235B-A22B-Instruct by 9.77 points and Qwen3.6-35B-A3B by 11.66 points, nearly matching the proprietary Seed 2.0 Pro captioner. This suggests that the claim-level reward optimization helps produce informative descriptions over extended temporal contexts. 

% \ours also substantially narrows the gap to proprietary captioners on VideoMME Overall and MMVU, while remaining competitive on MVBench and PerceptionTest. 
We observe remaining performance gaps on TOMATO, indicating that fine-grained motion understanding and temporal-relation reasoning are still challenging for caption-only transfer. Nevertheless, the overall pattern supports the goal of our reward objective: \ours improves captions as faithful proxies for downstream QA, rather than simply optimizing for caption-specific surface quality or evaluator preference.

\subsection{Ablation Study}
\label{subsec:ablation}
Table~\ref{tab:ablation_combined} ablates the reward design of \ours on DREAM-1K and VCapsBench. We include SFT as the no-RL baseline and compare GRPO variants that start from the same SFT checkpoint and use the same rollout and optimization setup. Holistic reward uses a whole-caption scalar reward without claim decomposition, testing whether response-level feedback is sufficient. The claim-level rows then isolate Base P/R in Eq.~\ref{eq:base-pr}, the precision-only and recall-only CuRe variants, and the full \ours reward in Eqs.~\ref{eq:cure-precision}--\ref{eq:cure-reward}. This setup separates whole-caption feedback from claim-level precision and recall credit; Figure~\ref{fig:ablation_pr} gives the corresponding precision--recall view.
\input{tables/table4_ablation}

\paragraph{Claim-level reward decomposition provides a more selective training signal.}
Holistic reward improves over SFT, raising DREAM-1K F1 from 34.67 to 37.42 and reducing VCapsBench IR from 16.81 to 15.65. This result indicates that response-level RL feedback is useful. Its gains are uneven, however: DREAM-1K recall decreases from 38.75 to 38.35, and VCapsBench CR decreases from 77.51 to 77.27. Full \ours changes this pattern. Compared with Holistic reward, it improves DREAM-1K precision by 11.68 points and F1 by 3.97 points, and improves all three VCapsBench metrics, with AR increasing by 2.59 points, IR decreasing by 1.87 points, and CR increasing by 1.33 points. The only trade-off is DREAM-1K recall, which drops by 2.09 points. This pattern suggests that whole-caption scalar feedback can improve the SFT warm start, but claim-level decomposition makes the reward more selective for faithful and informative captions.

\paragraph{Precision and recall calibration address complementary failure modes.}
Within claim-level RL, Base P/R provides the control without CuRe calibration. Precision-only training improves precision-side behavior relative to Base P/R, increasing DREAM-1K precision from 42.03 to 43.09 and reducing VCapsBench IR from 16.59 to 15.08. The same row loses DREAM-1K recall and VCapsBench CR, which indicates that factual selectivity alone does not preserve coverage. Recall-only training moves in the other direction: it reaches the highest DREAM-1K recall at 39.93 and raises VCapsBench CR to 78.48, but it remains weaker than full \ours on DREAM-1K precision and F1 and on VCapsBench AR and IR. Full \ours combines both sides, giving the best DREAM-1K precision and F1 and the best VCapsBench AR, IR, and CR. It does not maximize DREAM-1K recall, but it improves over Base P/R on recall while raising precision by 6.18 points and F1 by 2.61 points. These results indicate that factual faithfulness and descriptive coverage require different reward pressure, and the full CuRe reward balances them better than either single-side variant.

\subsection{Qualitative Analysis}
\label{subsec:qualitative-analysis}

Figure~\ref{fig:qualitative-main-case} shows the precision--coverage trade-off in one held-out DREAM-1K video. Both captions find the central sack interaction, so the difference is not basic scene recognition. The baseline turns the cartoon impact lines around the sack into unsupported causal content: water sprays out, and SpongeBob is being submerged or drenched. It also treats the later sack as the same sack that previously contained SpongeBob. These details fit a plausible story, but they are not directly supported by the sampled frames.

\ours describes the same event through visible relations. It names the burlap-like sack on the floor, the distorted reaction near the sack, and the later reaching action toward the sack. Rather than shortening the description, the caption changes the evidence type: inferred causes become verifiable visual claims. This matches the reward design in \S\ref{subsec:scoring}: unsupported claims receive no credit, while video-supported residual details can still receive bounded credit when they fall outside the reference caption. The case mirrors the pattern in Table~\ref{tab:video_hallusion_compact_results} and Table~\ref{tab:ablation_combined}: \ours raises factual precision without suppressing informative detail.
\input{figures/figure4_case_study}

%% file: tables/table1_hallusion_caption_bmks.tex
\begin{table*}[t]
\centering
% \caption{Caption and hallucination benchmark evaluation results with all scores normalized to percentages. 
% For EventHallusion, ``Entire”, ``Inter.”, and ``Mislead.” indicate rare-event recognition, temporal-switch hallucination, and prompt-misleading robustness subsets, respectively. 
% For VCapsBench, AR, IR, and CR denote Accuracy Rate, Inconsistency Rate (lower is better), and Coverage Rate, respectively. 
% For DREAM-1K, ``Prec.” and ``Rec.” stand for precision and recall. 
% \textbf{Bold} and \underline{underline} mark the best and second-best open-source results. Proprietary models are reference-only and excluded from highlighting.}
\caption{Evaluation on captioning and hallucination benchmarks. All scores are normalized as percentages. EventHallusion (Description): “Entire”, “Inter.”, and “Mislead.” denote rare-event recognition, temporal-switch hallucination, and prompt-misleading robustness. VCapsBench: AR, IR, and CR denote Accuracy Rate, Inconsistency Rate, and Coverage Rate, where lower is better for IR and higher is better otherwise. DREAM-1K: “Prec.” and “Rec.” denote precision and recall. \textbf{Bold} and \underline{underline} mark the best and second-best open-source results; proprietary models are reference-only and excluded from highlighting.}
\label{tab:video_hallusion_compact_results}
\setlength{\tabcolsep}{4.0pt}
\renewcommand{\arraystretch}{1.08}
\scriptsize
\resizebox{\textwidth}{!}{
\begin{tabular}{l|cccccccccc}
\toprule
\multirow{2}{*}{Model}
& \multicolumn{4}{c}{EventHallusion (Description)}
& \multicolumn{3}{c}{VCapsBench}
& \multicolumn{3}{c}{DREAM-1K} \\
\cline{2-5}\cline{6-8}\cline{9-11}
& Entire$\uparrow$ & Inter.$\uparrow$ & Mislead.$\uparrow$ & Overall$\uparrow$
& AR$\uparrow$
& IR$\downarrow$
& CR$\uparrow$
& Prec.$\uparrow$
& Rec.$\uparrow$
& F1$\uparrow$ \\
\midrule

\rowcolor{gray!15}
\multicolumn{11}{c}{\emph{Proprietary models}} \\
\midrule
Seed 2.0 Pro~\citep{bytedance2026seed2}
& 64.22 & 76.17 & 96.84 & 77.83
& 60.37 & 14.59 & 70.68
& 32.97 & 50.80 & 39.99 \\
Gemini 3.0 Pro~\citep{google2025gemini}
& 65.13 & 68.39 & 98.04 & 74.59
& 70.29 & 15.44 & 83.13
& 42.67 & 39.26 & 40.89 \\
\midrule

\rowcolor{gray!15}
\multicolumn{11}{c}{\emph{Open-source models}} \\
\midrule
LLaVA-Video-7B~\citep{Zhang2024LLaVAVideo}
& 27.50 & 32.60 & 75.50 & 41.40
& 45.11 & 17.46 & 54.65
& 37.90 & 28.40 & 32.50 \\
OwlCap-7B~\citep{Zhong2025OwlCap}
& 20.18 & 51.81 & 80.00 & 49.87
& 43.82 & 15.79 & 52.04
& 34.10 & 35.30 & 34.70 \\
Tarsier2-7B~\citep{Yuan2025Tarsier2}
& \textbf{54.60} & 53.10 & \underline{93.70} & \underline{63.30}
& 38.51 & \underline{13.89} & 44.73
& \underline{41.50} & \underline{38.80} & \underline{40.10} \\
InternVL3.5-30B-A3B~\citep{wang2025internvl3}
& 34.86 & 54.40 & 75.79 & 54.16
& 56.85 & 17.43 & 68.86
& 39.07 & 28.17 & 32.74 \\
Qwen3-VL-30B-A3B-Instruct~\citep{QwenTeam2025Qwen3VL}
& 32.11 & 60.10 & 87.37 & 58.94
& 61.26 & 14.30 & 71.49
& 33.35 & 38.32 & 35.66 \\
Tarsier-34B~\citep{Wang2024Tarsier}
& 38.50	& 40.40 & 83.20 & 50.10
& - & - & -
& 41.40 & 32.40 & 36.30 \\
Qwen2.5-VL-72B-Instruct~\citep{tongyi2025qwen25vl}
& 24.77	& 62.69 & 81.05 & 56.68
& 61.20 & 16.02 & 67.81
& 34.72 & 28.90 & 31.54 \\
Qwen3-VL-235B-A22B-Instruct~\citep{QwenTeam2025Qwen3VL}
& 39.45 & \underline{65.28} & 81.05 & 61.96
& \underline{64.80} & 16.09 & \underline{77.23}
& 29.48 & \textbf{40.77} & 34.22 \\
\midrule
\ours
& \underline{45.87} & \textbf{69.95} & \textbf{95.79} & \textbf{69.52}
& \textbf{67.77} & \textbf{13.78} & \textbf{78.60}
& \textbf{48.21} & 36.26 & \textbf{41.39} \\
\bottomrule
\end{tabular}
}
\end{table*}

%% file: tables/table2_downstream_training.tex
\begin{table*}[t]
\centering
% \caption{Downstream pretraining utility of video caption datasets with varying source videos and annotating models. For each corpus, ``\ours-Recap.” replaces the original captions with \ours-generated captions for Stage-2 VLM training. The training recipe of other stages, as well as the evaluation protocol are held fixed across experiments. Scores are normalized in percentages, and higher is better for all metrics. \textbf{Bold} marks the better result within each paired comparison.}
\caption{Evaluation of re-annotating Stage-2 captions with \ours on downstream performance across different data sources. “CuRe-Recap.” denotes Stage-2 training with \ours-reannotated captions. Other training stages and the evaluation protocol are kept fixed. \textbf{Bold} marks the better result in each paired comparison.}
\label{tab:downstream_pretraining_utility}
\setlength{\tabcolsep}{3.6pt}
\renewcommand{\arraystretch}{1.06}
\scriptsize
\resizebox{\textwidth}{!}{%
\begin{tabular}{llccccc|cccc}
\toprule
\multirow{2}{*}{Stage-2 Data Source}
& \multirow{2}{*}{Caption Source}
& \multirow{2}{*}{MMVU}
& \multirow{2}{*}{MVBench}
& \multirow{2}{*}{MotionBench}
& \multirow{2}{*}{PerceptionTest}
& \multirow{2}{*}{TOMATO}
& \multicolumn{4}{c}{VideoMME} \\
\cmidrule(lr){8-11}
& & & & & & 
& Overall & Long & Medium & Short \\
\midrule

% \rowcolor{gray!12}
% \multicolumn{11}{c}{\emph{Qwen3 4B + Qwen3 ViT}} \\
% \midrule

\multirow{2}{*}{Molmo2-Cap (104k)}
& Original
& \textbf{38.10} & 45.10 & 47.98 & 48.99 & 22.64
& 52.15 & 42.33 & 49.33 & 64.78 \\
& \ours{} Recap.
& 37.40 & \textbf{48.83} & \textbf{49.38} & \textbf{52.30} & \textbf{26.08}
& \textbf{54.07} & \textbf{44.22} & \textbf{50.89} & \textbf{67.11} \\
\midrule

\multirow{2}{*}{Tarsier2-Recap (585k)}
& Original
& 25.40 & 48.85 & 49.00 & 51.84 & 25.00
& 51.52 & 40.78 & 48.78 & 65.00 \\
& \ours{} Recap.
& \textbf{28.70} & \textbf{49.48} & \textbf{50.42} & \textbf{53.15} & \textbf{26.68}
& \textbf{53.70} & \textbf{42.56} & \textbf{51.67} & \textbf{66.89} \\
\midrule

\multirow{2}{*}{LLaVA-Video-Caption (178k)}
& Original
& 31.20 & \textbf{49.08} & 48.86 & 52.68 & 22.57
& 52.30 & 39.00 & 51.56 & 66.33 \\
& \ours{} Recap.
& \textbf{37.30} & 48.73 & \textbf{50.32} & \textbf{53.25} & \textbf{24.93}
& \textbf{54.37} & \textbf{42.22} & \textbf{52.00} & \textbf{68.89} \\
\bottomrule
\end{tabular}%
}
\end{table*}

%% file: tables/table3_prism.tex
\begin{table*}[t]
\centering
% \caption{VQA utility of different video captioners under the Prism protocol. Seed 2.0 Pro is used as the fixed QA model to answer benchmark questions based on captioner-generated video description. Scores are normalized in percentages and higher is better. \textbf{Bold} and \underline{underlining} mark the best and second-best open-source results, respectively. Proprietary models are listed as reference only and excluded from highlighting.}
\caption{Caption-to-QA evaluation of different video captioners under the Prism protocol. Seed 2.0 Pro~\citep{bytedance2026seed2} answers benchmark questions using only the video description generated by each captioner. \textbf{Bold} and \underline{underline} mark the best and second-best open-source results, respectively.}
\label{tab:prism_model_benchmark_results}
\setlength{\tabcolsep}{3.4pt}
\renewcommand{\arraystretch}{1.06}
\scriptsize
\resizebox{\textwidth}{!}{%
\begin{tabular}{lccccccccccc}
\toprule
Captioner
& \multicolumn{4}{c}{VideoMME}
& \multicolumn{3}{c}{MMVU}
& \multicolumn{1}{c}{MVBench}
& \multicolumn{1}{c}{MotionBench}
& \multicolumn{1}{c}{TOMATO}
& \multicolumn{1}{c}{PerceptionTest} \\
\cmidrule(lr){2-5}
\cmidrule(lr){6-8}
\cmidrule(lr){9-9}
\cmidrule(lr){10-10}
\cmidrule(lr){11-11}
\cmidrule(lr){12-12}
& Overall & Short & Medium & Long
& MCQ & Open & Overall
& Overall & Overall & Score & Overall \\
\midrule

\rowcolor{gray!12}
\multicolumn{12}{c}{\emph{Proprietary captioners}} \\
\midrule
Seed 2.0 Pro
& 66.63 & 74.67 & 68.00 & 57.22
& 78.24 & 55.73 & 69.80
& 56.23 & 53.19 & 35.58 & 65.75 \\
Gemini 3.0 Pro
& 68.52 & 75.11 & 69.67 & 60.78
& 77.28 & 58.13 & 70.10
& 59.13 & 50.85 & 35.31 & 63.73 \\

\midrule
\rowcolor{gray!12}
\multicolumn{12}{c}{\emph{Open-source captioners}} \\
\midrule
Tarsier2-7B
& 33.19 & 40.44 & 30.78 & 28.33
& 73.12 & 42.93 & 61.80
& 35.70 & 33.00 & 11.39 & 37.51 \\
OwlCap-7B
& 44.56 & 53.78 & 41.56 & 38.33
& 67.84 & 44.00 & 58.90
& 45.10 & 37.01 & 21.77 & 52.58 \\
Qwen3-VL-30B-A3B-Instruct
& 52.04 & 61.00 & 51.33 & 43.78
& 72.48 & 45.87 & 62.50
& 51.32 & 41.02 & 21.50 & 60.05 \\
Qwen3.6-35B-A3B
& \underline{61.15} & \underline{73.11} & \underline{65.67} & 44.67
& \underline{75.04} & \underline{49.87} & \underline{65.60}
& \underline{53.60} & \underline{45.02} & 21.90 & \underline{61.72} \\
Qwen3-VL-235B-A22B-Instruct
& 53.81 & 61.33 & 53.56 & \underline{46.56}
& 73.12 & 48.53 & 63.90
& 47.80 & 43.18 & \textbf{23.05} & 61.26 \\

\midrule
\ours
& \textbf{65.44} & \textbf{73.67} & \textbf{66.33} & \textbf{56.33}
& \textbf{77.44} & \textbf{51.73} & \textbf{67.80}
& \textbf{54.43} & \textbf{47.86} & \underline{22.91} & \textbf{63.56} \\
\bottomrule
\end{tabular}%
}
\end{table*}

%% file: tables/table4_ablation.tex
\begin{table*}[t]
\centering
\caption{Ablation of training strategy and RL reward design on DREAM-1K and VCapsBench. SFT is the no-RL baseline; all SFT + GRPO rows start from the same SFT checkpoint and use the same GRPO setup, differing only in the reward. \textbf{Bold} marks the best value in each metric.}
\label{tab:ablation_combined}
\setlength{\tabcolsep}{3.8pt}
\renewcommand{\arraystretch}{1.08}
\scriptsize
\resizebox{0.9\textwidth}{!}{%
\begin{tabular}{@{}lllcccccc@{}}
\toprule
\multirow{2}{*}{Method}
& \multirow{2}{*}{Training}
& \multirow{2}{*}{RL reward}
& \multicolumn{3}{c}{DREAM-1K}
& \multicolumn{3}{c}{VCapsBench} \\
\cmidrule(lr){4-6}\cmidrule(lr){7-9}
& & & P$\uparrow$ & R$\uparrow$ & F1$\uparrow$
& AR$\uparrow$ & IR$\downarrow$ & CR$\uparrow$ \\
\midrule
SFT
& SFT only & --
& 31.37 & 38.75 & 34.67
& 64.48 & 16.81 & 77.51 \\
Holistic reward
& SFT + GRPO & Whole-caption
& 36.53 & 38.35 & 37.42
& 65.18 & 15.65 & 77.27 \\
Base P/R
& SFT + GRPO & Claim P/R
& 42.03 & 35.99 & 38.78
& 65.08 & 16.59 & 78.03 \\
Precision-only
& SFT + GRPO & Claim precision
& 43.09 & 34.59 & 38.38
& 66.03 & 15.08 & 77.76 \\
Recall-only
& SFT + GRPO & Claim recall
& 36.35 & \textbf{39.93} & 38.06
& 65.69 & 16.30 & 78.48 \\
\rowcolor{gray!12}
\textbf{\ours}
& SFT + GRPO & Claim precision + recall
& \textbf{48.21} & 36.26 & \textbf{41.39}
& \textbf{67.77} & \textbf{13.78} & \textbf{78.60} \\
\bottomrule
\end{tabular}%
}
\end{table*}

\begin{figure}[t]
\centering
\includegraphics[width=0.92\textwidth]{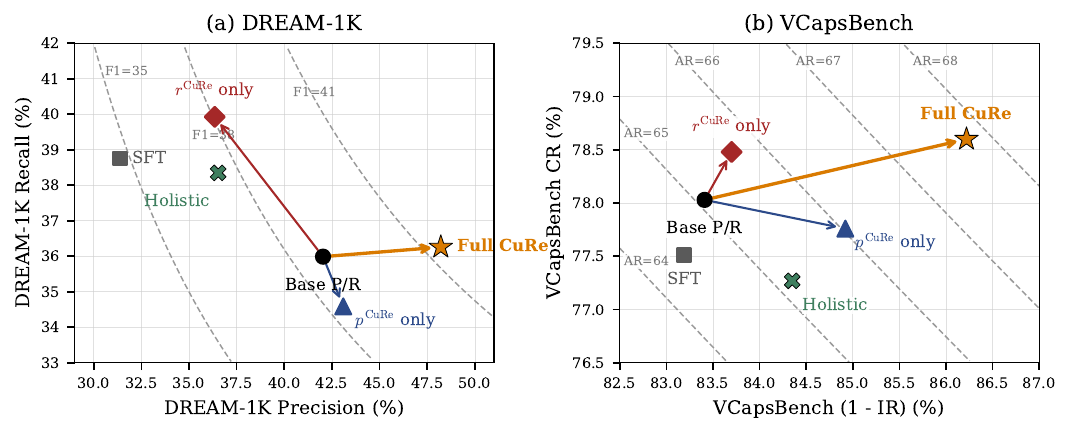}
\caption{Ablation variants in DREAM-1K precision--recall space and VCapsBench $(1-\mathrm{IR})$--CR space. SFT, Holistic reward, and full \ours compare the training signal, while Base P/R and the precision-only and recall-only rows isolate claim-level reward components.}
\label{fig:ablation_pr}
\end{figure}

%% file: figures/figure4_case_study.tex
\begin{figure}[t]
\centering
\captionsetup{skip=8pt}
\includegraphics[width=\linewidth]{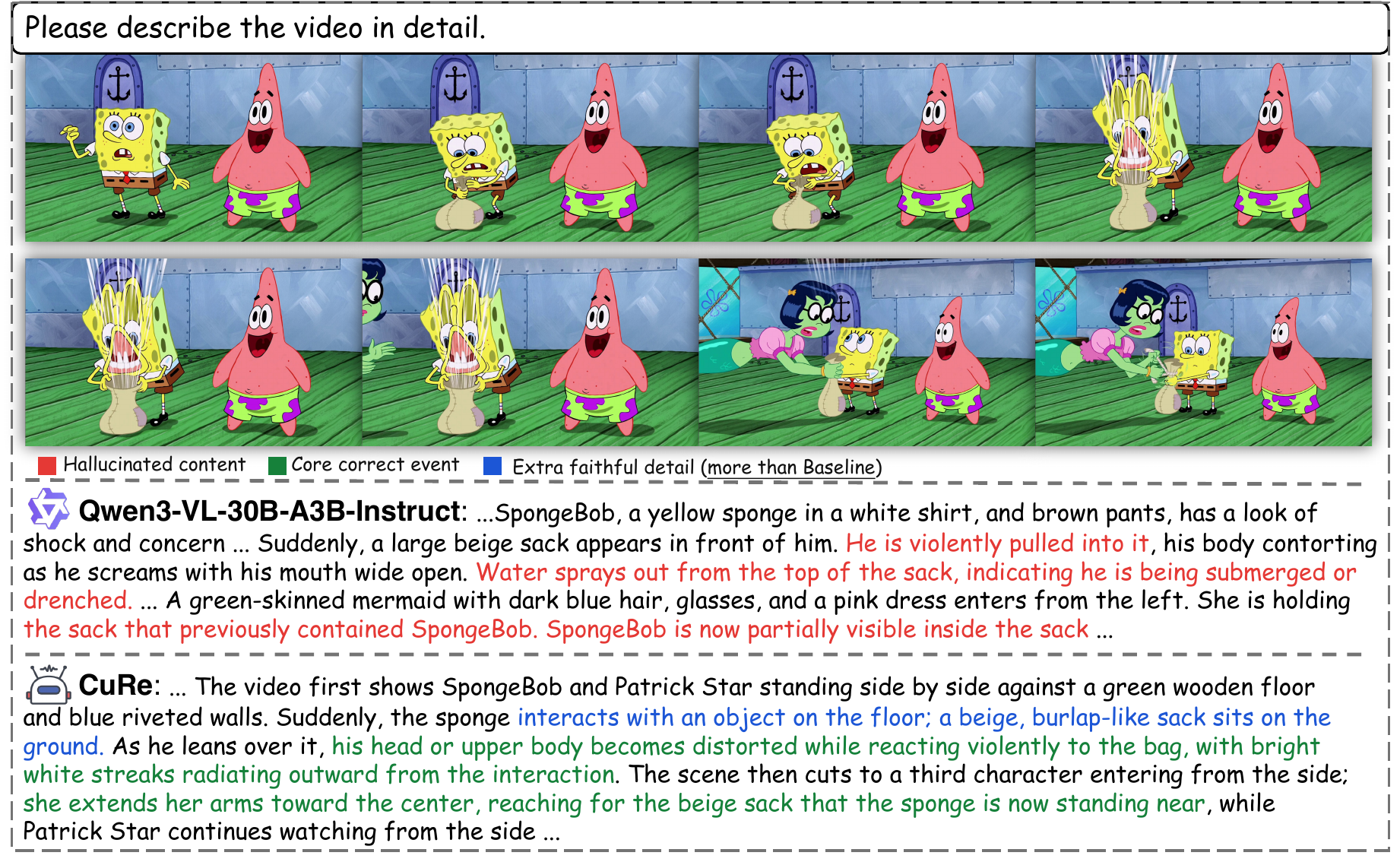}
\caption{Qualitative comparison on a held-out DREAM-1K video. Qwen3-VL-30B-A3B-Instruct recognizes the sack interaction but adds unsupported causal details, such as water spraying out and SpongeBob being submerged or still inside the sack. \ours keeps the shared event while adding visible details about the sack, impact streaks, and the later reaching action. \textcolor{red}{Red} marks hallucinated content, \textcolor{green!45!black}{green} marks the core correct event, and \textcolor{blue}{blue} marks faithful details added by \ours. Additional examples are provided in Appendix~\ref{app:qualitative}.}
\label{fig:qualitative-main-case}
\end{figure}

%% file: sections/05_conclusion.tex
%% ============================================================================
%% 5. CONCLUSION  (rendered as Section 5 after the Conclusion / Limitations swap)
%% ============================================================================
\section{Conclusion}
\label{sec:conclusion}

We presented \ours, a claim-level rubric reward whose reference-anchored calibration mechanism induces teacher-anchored mode seeking at the reward layer. Empirically, this scoring scheme yields captions that simultaneously gain factual faithfulness and descriptive density over supervised and reference-bound RL baselines. The resulting captioner provides stronger pretraining supervision and, under caption-to-QA evaluation, serves as a video proxy that outperforms much larger open-source captioners. The pattern across these three families suggests that claim-level rubric design offers a viable alternative to whole-caption rewards and may extend to other open-ended generation tasks. While the calibration mechanism currently uses a single reference-caption source, the framework supports multi-source or ensemble references, and stronger treatment of partial references remains a direction for future work.

%% file: sections/06_discussion_limitations.tex
%% ============================================================================
%% 6. LIMITATIONS  (rendered as Section 6 after the Conclusion / Limitations swap)
%% ============================================================================

%\newpage 

\section{Limitations}
\label{sec:limitations}

Despite the empirical gains of \ours, our current instantiation remains compute-intensive. The reward stack calls a large VLM for claim decomposition, verification, and matching during training, which limits the breadth of verifier variants, reference-caption sources, and multi-run evaluations we can afford. This cost is a limitation of the present implementation rather than of the claim-level reward interface itself; developing cheaper or more specialized verifiers is an important direction for future work.

Our empirical scope also bounds the claims. The experiments cover Caption Quality Evaluation, CuRe-Annotated Caption for VLM Pretraining, and Prism Evaluation, but each setting still fixes some design choices: the main policy uses one backbone, the pretraining study uses one downstream learner, and Prism uses one fixed QA model. The consistent gains across these settings suggest that claim-level rubric rewards transfer within the evaluated regimes, but they do not establish universality across all backbones, verifiers, answerers, or video domains.

\section{Ethical Considerations}
\label{sec:ethical-considerations}

This work improves dense video captioning, which can turn visual content into searchable text. This is useful for video-language research and data engineering, but it may also increase privacy risks when videos contain people, locations, private settings, or visible text. The captioner is not designed or evaluated for person identification, tracking, surveillance, or high-stakes decision making. Applications involving people or sensitive scenes should therefore add privacy review, content filtering, and human oversight.

Although \ours aims to make dense captions more faithful and informative, generated captions remain model-produced annotations rather than ground truth. They may still hallucinate, omit context, or reflect biases in the source videos, reference-caption generators, and evaluators. When recaptions are reused for training or evaluation, such errors can propagate into downstream models or benchmarks. Benchmark gains should therefore not be read as evidence that \ours behaves uniformly across all communities, languages, activities, or recording conditions. If artifacts derived from this work are released, they should include clear intended-use guidance and restrictions against surveillance and high-stakes automated decision-making.

%% file: sections/appendix.tex
\appendix

%% ==========================================================================
%% A. POST-TRAINING AND REWARD RUNTIME
%% ==========================================================================
\section{Post-Training and Reward Runtime}
\label{app:training_config}

\S\ref{subsec:setup} describes the two-stage post-training pipeline at a high level. Table~\ref{tab:training_config} records the full configuration needed to reproduce each stage, and Table~\ref{tab:reward_runtime} specifies the reward runtime that drives GRPO training.

\input{tables/appendix/tab_training_config}

\input{tables/appendix/tab_reward_runtime}

\paragraph{Length-control regularizer.}
At reward time, we compute the generated-to-reference claim ratio $\rho = |\mathcal{C}| / |\mathcal{C}^{\star}|$ and subtract a bounded penalty
\begin{equation}
\Omega = \eta \cdot \min\!\biggl(1,\;
\frac{\max(0,\, \rho - \tau)}{\kappa}
\biggr),
\label{eq:overgen-regularizer}
\end{equation}
where $\tau$ is the tolerance threshold, $\kappa$ controls the penalty growth rate, and $\eta$ is the maximum penalty. This saturated penalty suppresses verbose captions with excessively low-credit or redundant claims while maintaining stable optimization.

\paragraph{Hyperparameter selection.}
All hyperparameters were selected based on preliminary experiments on a held-out subset. No systematic search (grid, random, or Bayesian) was performed due to the cost of online 235B reward calls. The category priors $\omega_k$ in Appendix~\ref{app:priors} reflect domain knowledge about which visual dimensions dense captions most often underdescribe. All main results and ablations are single training runs; we do not report standard deviations.

%% ==========================================================================
%% B. CLAIM SCHEMA AND CATEGORY PRIORS
%% ==========================================================================
\section{Claim Schema and Category Priors}
\label{app:priors}

All claim-level operations use the same ten visual categories. The category names serve as labels for decomposition, verification, matching, and aggregation; the weights are priors for aggregating per-category precision and recall against reference claims.

\input{tables/appendix/tab_category_priors}

Action \& Interaction receives the highest prior ($\omega_k = 1.8$) because dense video captions most often underdescribe actions and interactions. Static attribute categories receive the baseline weight of $1.0$.

%% ==========================================================================
%% C. EVALUATION AND REPORTING PROTOCOLS
%% ==========================================================================
\section{Evaluation and Reporting Protocols}
\label{app:eval-protocols}

This section details the evaluation protocols corresponding to the experimental subsections in the main text. \S\ref{app:eval-caption} covers caption quality evaluation (cf.\ \S\ref{subsec:open-source-caption-benchmarks}), \S\ref{app:eval-prism} covers Prism evaluation (cf.\ \S\ref{subsec:prism-caption-to-qa}), and \S\ref{app:eval-ablation} covers the training and reward-design ablation (cf.\ \S\ref{subsec:ablation}). The downstream pretraining protocol is documented separately in \S\ref{app:downstream-training}.

%% --- C.1 Caption Quality Evaluation ---
\subsection{Caption Quality Evaluation}
\label{app:eval-caption}

Table~\ref{tab:direct_eval_protocol} records the three direct caption evaluations used in Table~\ref{tab:video_hallusion_compact_results}. All entries are reported as percentages. Proprietary systems are included as reference upper bounds and are excluded from open-source bolding.

\input{tables/appendix/tab_direct_eval_protocol}

Table~\ref{tab:baseline_groups} documents baseline grouping and result sourcing.

\input{tables/appendix/tab_baseline_groups}

%% --- C.2 Prism Evaluation ---
\subsection{Prism Evaluation}
\label{app:eval-prism}

The Prism-style protocol evaluates whether a caption preserves enough information for a text-only QA model to answer benchmark questions. Table~\ref{tab:prism_protocol} separates the two stages.

\input{tables/appendix/tab_prism_protocol}

The reported benchmarks are VideoMME, MMVU, MVBench, MotionBench, TOMATO, and PerceptionTest. VideoMME reports Overall plus Short, Medium, and Long splits; MMVU reports MCQ, Open, and Overall; the remaining benchmarks report Overall.

%% --- C.3 Reward Granularity and Reward-Component Ablation ---
\subsection{Reward Granularity and Reward-Component Ablation}
\label{app:eval-ablation}

Table~\ref{tab:ablation_protocol} defines the rows in Table~\ref{tab:ablation_combined}. All RL rows start from the same SFT initialization and use the same rollout setup; only the reward changes.

\input{tables/appendix/tab_ablation_protocol}

%% ==========================================================================
%% D. CURE-ANNOTATED CAPTION FOR VLM PRETRAINING
%% ==========================================================================
\section{CuRe-Annotated Caption for VLM Pretraining}
\label{app:downstream-training}

This section gives the training details for the downstream pretraining utility study in Table~\ref{tab:downstream_pretraining_utility}. Following the ShareGPT4V~\citep{Chen2024ShareGPT4V,Chen2024ShareGPT4Video} and CapRL~\citep{Xing2026CapRL} pretraining paradigm, we use three stages: initial visual-language alignment, caption-data further pretraining, and instruction SFT. The study is designed as a controlled data-quality test: within each paired row, the model architecture, Stage~1 alignment, Stage~3 SFT, and evaluation suite are fixed, and only the Stage~2 caption source changes.

\paragraph{Model architecture.}
The downstream learner uses a Qwen3-VL-4B-Instruct vision encoder, a Qwen3-4B Base language backbone, and a randomly re-initialized multimodal projector (Kaiming normal initialization). The vision encoder is kept pretrained, we load base weights rather than instruction-tuned weights to avoid instruction-tuning bias, and the projector is left to be aligned in Stage~1.

\paragraph{Training setting.}
Stage~1 performs initial alignment on BLIP-3-Kale image-caption data. Only the projector is trainable while the ViT and LLM are frozen. Stage~2 performs further pretraining on video-caption data with all modules unfrozen. For each Stage~2 corpus, we train on either the original captions or on \ours-generated recaptions. The reported paired corpora are Molmo2-Cap, Tarsier2-Recap, and LLaVA-Video-Caption. Stage~3 performs VideoQA instruction SFT starting from the corresponding Stage~2 checkpoint, using the same merged VideoQA data (Molmo2-AskModelAnything 20K + LLaVA-Video-178K Stage-3 80K) for every paired comparison.

\paragraph{Controlled comparisons.}
Table~\ref{tab:downstream_protocol} lists the hyperparameters for the three stages. The paired comparisons in Table~\ref{tab:downstream_pretraining_utility} keep these settings fixed within each corpus pair. We report MMVU, MVBench, MotionBench, PerceptionTest, TOMATO, and VideoMME accuracies. When computing the average gain, we average benchmark-level overall columns; the VideoMME duration splits are diagnostic and are not counted as separate benchmarks.

\input{tables/appendix/tab_downstream_protocol}

%% ==========================================================================
%% E. QUALITATIVE ANALYSIS
%% ==========================================================================
\clearpage
\section{Qualitative Analysis}
\label{app:qualitative}

Figure~\ref{fig:qualitative-main-case} and Appendix Cases~1--2 show three representative side-by-side caption comparisons on held-out DREAM-1K videos. Figure~\ref{fig:qualitative-main-case} compares \ours against its same-size baseline (Qwen3-VL-30B-A3B-Instruct); Appendix Cases~1--2 compare \ours against the much larger Qwen3-VL-235B-A22B-Instruct. Captions are excerpts from each model's output; ``\dots'' indicates omitted unrelated text. \textcolor{red}{Red} marks hallucinated or unsupported baseline content; \textcolor{blue}{blue} marks visually supported details that \ours adds.

\begin{center}
\begin{minipage}{\textwidth}
\centering
\includegraphics[width=\textwidth]{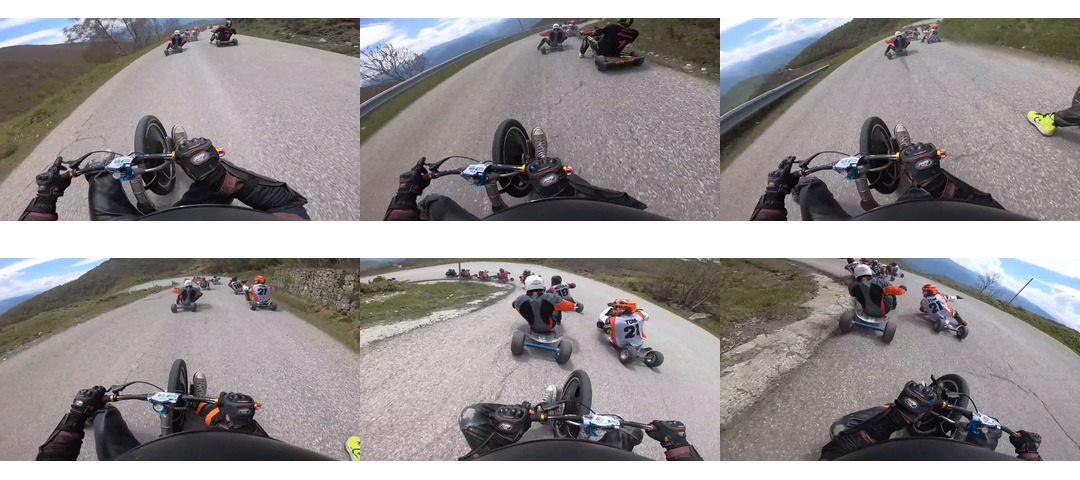}
\vspace{1mm}

\begin{captioncasebox}{Qwen3-VL-235B-A22B-Instruct baseline}
... The video captures an exhilarating downhill race on a winding mountain road, filmed from the first-person perspective of a rider on a gravity-powered vehicle - \hall{likely a street luge or downhill longboard}. ... Ahead, a group of competitors races down the same road, \hall{each on similar low-profile boards}. They wear colorful racing suits and helmets ...
\end{captioncasebox}
\vspace{1mm}
\begin{captioncasebox}{\ours}
... This video presents \cureextra{a first-person POV} from a rider in a downhill race on a paved mountain road. The rider's hands grip handlebars, and a large front wheel is partially visible ... Ahead of the POV rider, \cureextra{a group of other competitors is navigating the descent}. These participants are riding \cureextra{specialized low-profile vehicles}, with riders adjusting their legs through the bend. The camera wearer closes the gap, overtakes from the left, and prepares for the next left turn.
\end{captioncasebox}
\begingroup
\captionsetup{hypcap=false}
\captionof{figure}{Appendix Case 1: qualitative caption comparison on a held-out DREAM-1K video. Blue marks visually supported details added by \ours; red marks unsupported or incorrect baseline content.}
\label{fig:qwen-base-case-02}
\endgroup
\end{minipage}
\end{center}

\begin{center}
\begin{minipage}{\textwidth}
\centering
\includegraphics[width=\textwidth]{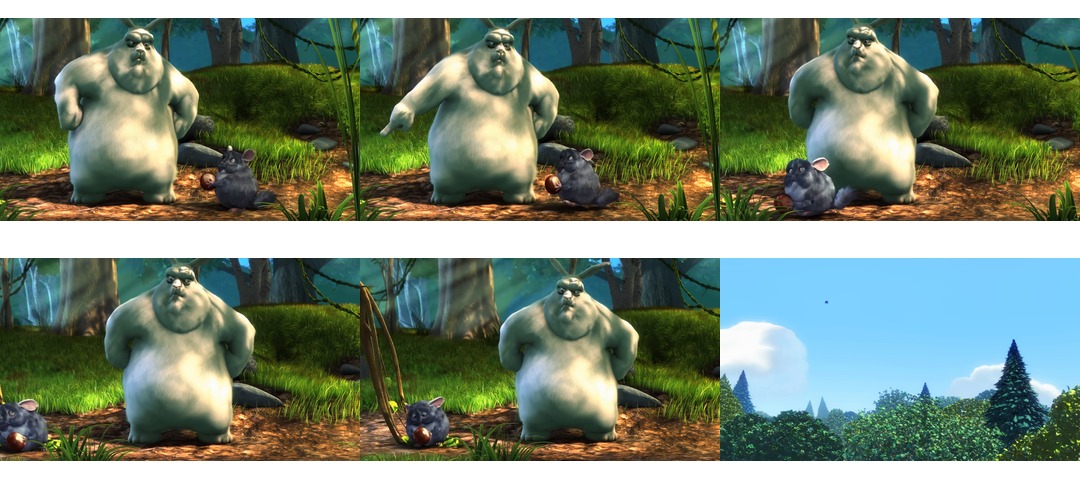}
\vspace{1mm}

\begin{captioncasebox}{Qwen3-VL-235B-A22B-Instruct baseline}
... In a lush, sun-dappled forest clearing, two animated creatures stand on a dirt path ... The larger creature is a massive, muscular white rabbit ... Beside it, a much smaller, dark gray mouse holds a shiny acorn in its tiny paws. ... The mouse, \hall{undeterred}, continues to hold the acorn and looks around,  \hall{contemplating the rabbit's words} ...
\end{captioncasebox}
\vspace{1mm}
\begin{captioncasebox}{\ours}
... The video presents a 3D animated forest scene with a giant rabbit figure and a small dark-furred creature. The small creature \cureextra{holds a round reddish-brown nut and reacts to the rabbit's direction}. As the sequence progresses, the large creature \cureextra{extends an arm or points}, and the small creature moves toward the green bush. The bush is then \cureextra{revealed as a disguised wooden trap}, turning the exchange into a directed trap setup rather than a simple forest conversation.
\end{captioncasebox}
\begingroup
\captionsetup{hypcap=false}
\captionof{figure}{Appendix Case 2: qualitative caption comparison on a held-out DREAM-1K video. Blue marks visually supported details added by \ours; red marks unsupported or incorrect baseline content.}
\label{fig:qwen-base-case-03}
\endgroup
\end{minipage}
\end{center}

%% ==========================================================================
%% F. PROMPT TEMPLATES
%% ==========================================================================
\section{Prompt Templates}
\label{app:prompts}

This section lists the prompt contracts used in reward training and evaluation. Prompts are grouped into reward-training prompts (\S\ref{app:reward-prompts}) used by the online reward runtime, and evaluation prompts (\S\ref{app:eval-prompts-section}) used for direct caption and Prism-style evaluation.

\subsection{Reward Training Prompts}
\label{app:reward-prompts}

The reward prompts reproduce the English templates used by the claim decomposer, video verifier, and matcher roles in Table~\ref{tab:reward_runtime}. Placeholders such as \texttt{\{caption\}} denote values filled at runtime. Figure~\ref{fig:prompt-decompose} shows the shared claim-schema decomposition prompt used for rollout claims and precomputed Gemini reference claims; Figure~\ref{fig:prompt-verify} shows video-only claim verification; Figure~\ref{fig:prompt-match} shows the per-category one-to-one matcher.

\begin{center}
\fbox{\begin{minipage}{0.94\textwidth}
\input{prompt_templates/reward_decompose_en_box.tex}
\end{minipage}}
\begingroup
\captionsetup{hypcap=false}
\captionof{figure}{Full prompt template for claim-schema decomposition in reward training.}
\label{fig:prompt-decompose}
\endgroup
\end{center}

\begin{center}
\fbox{\begin{minipage}{0.94\textwidth}
\input{prompt_templates/reward_verify_en_box.tex}
\end{minipage}}
\begingroup
\captionsetup{hypcap=false}
\captionof{figure}{Full prompt template for video-only claim verification.}
\label{fig:prompt-verify}
\endgroup
\end{center}

\begin{center}
\fbox{\begin{minipage}{0.94\textwidth}
\input{prompt_templates/reward_category_match_en_one_to_one_box.tex}
\end{minipage}}
\begingroup
\captionsetup{hypcap=false}
\captionof{figure}{Full prompt template for per-category one-to-one semantic matching.}
\label{fig:prompt-match}
\endgroup
\end{center}

\subsection{Evaluation Prompts}
\label{app:eval-prompts-section}

The evaluation prompts are short and shared across direct caption evaluation and Prism-style caption-to-QA. Figure~\ref{fig:prompt-direct-caption} shows the caption-only prompt used for direct caption evaluation and Prism Stage~1; Figure~\ref{fig:prompt-prism-qa} shows the caption-conditioned QA prompt used by the fixed Stage~2 answerer.

\begin{center}
\fbox{\begin{minipage}{0.94\textwidth}
\small
\textbf{Caption generation.}
Provide a detailed and comprehensive description of the video. The description should be sufficiently detailed and rich.
\end{minipage}}
\begingroup
\captionsetup{hypcap=false}
\captionof{figure}{Caption-only prompt used for direct caption evaluation and Prism Stage~1.}
\label{fig:prompt-direct-caption}
\endgroup
\end{center}

\begin{center}
\fbox{\begin{minipage}{0.94\textwidth}
\small
\textbf{Prism Stage 2: caption-conditioned QA.}
Given the video caption below, carefully analyze the content and answer the multiple-choice question. The prompt provides the generated caption, the benchmark question, and the answer options. The QA model must answer directly with the option letter from the given choices.
\end{minipage}}
\begingroup
\captionsetup{hypcap=false}
\captionof{figure}{Stage~2 prompt used by the fixed QA model in Prism-style evaluation.}
\label{fig:prompt-prism-qa}
\endgroup
\end{center}

%% ==========================================================================
%% H. DATA PRIVACY AND CONTENT SCREENING
%% ==========================================================================
\section{Data Privacy and Content Screening}
\label{app:privacy}

\paragraph{Public benchmarks.}
EventHallusion, VCapsBench, DREAM-1K, VideoMME, MMVU, MVBench, MotionBench, TOMATO, and PerceptionTest are publicly released research benchmarks distributed under their respective terms. We use them without modification and do not perform additional screening beyond what their creators applied.

\paragraph{Training videos.}
All training videos are sourced from publicly available datasets and web platforms that permit research use. We did not collect new video data from human subjects. The videos are used solely for model training and evaluation; we do not attempt to identify, track, or profile individuals appearing in the videos.

%% file: tables/appendix/tab_training_config.tex
\begin{table*}[htbp]
\centering
\small
\setlength{\tabcolsep}{4pt}
\renewcommand{\arraystretch}{1.12}
\caption{Policy post-training configuration for the main \ours model.}
\begin{tabular}{p{0.14\textwidth}p{0.22\textwidth}p{0.56\textwidth}}
\toprule
Stage & Parameter & Value \\
\midrule
\multicolumn{3}{l}{\emph{Supervised Fine-Tuning (SFT)}} \\
& Data & 78{,}144 Gemini-3.0-Pro distilled video caption samples \\
& Optimization & 1 epoch; peak LR $4{\times}10^{-6}$; cosine decay; AdamW ($\beta_1{=}0.9$, $\beta_2{=}0.999$) \\
& Video budget & 2 fps; max 512 frames; dynamic resolution (max 1{,}024 visual tokens per video) \\
\midrule
\multicolumn{3}{l}{\emph{GRPO Reinforcement Learning}} \\
& Data & Disjoint 8{,}000-video subset (no overlap with SFT or evaluation) \\
& Rollout group size & $G=8$ captions per video \\
& Rollout sampling & Temperature 1.0; top-$p$ = 1.0; max 8{,}192 response tokens \\
& Actor LR / schedule & $2{\times}10^{-6}$ $\to$ $2{\times}10^{-7}$; cosine decay; 5 warmup steps \\
& Batch size & Train 64, mini-batch 64, per-GPU micro-batch 1 \\
& Sequence limits & 24{,}576 prompt tokens; 8{,}192 response tokens \\
& KL control & Low-variance KL loss; $\beta = 0.005$; not added to reward \\
& Video budget & 2 fps; max 256 frames; dynamic resolution \\
& Training duration & 125 steps; $\approx$62.5 hours on 32$\times$H20 ($\approx$2{,}000 GPU-hours) \\
\midrule
\multicolumn{3}{l}{\emph{Infrastructure}} \\
& Hardware & 32 NVIDIA H20 GPUs \\
& Parallelism (actor) & TP=4, CP=2, EP=4, PP=1 (Megatron) \\
& Parallelism (rollout) & TP=4 (vLLM); GPU memory utilization 0.55 \\
& Framework & verl 0.7.0~\citep{sheng2024verl}; vLLM 0.11.0~\citep{kwon2023vllm} \\
& Software & PyTorch 2.8.0; Transformers 4.57.3; Python 3.12 \\
\midrule
\multicolumn{3}{l}{\emph{Data Integrity}} \\
& Leakage control & SFT and GRPO videos are excluded from all evaluation benchmarks \\
\bottomrule
\end{tabular}
\label{tab:training_config}
\end{table*}

%% file: tables/appendix/tab_reward_runtime.tex
\begin{table*}[htbp]
\centering
\scriptsize
\setlength{\tabcolsep}{3pt}
\renewcommand{\arraystretch}{1.12}
\caption{Reward runtime configuration. The Qwen3-VL-235B roles use top-$p$ = 1.0 (vLLM default). Gemini calls use Google Vertex API defaults for temperature and top-$p$. All roles receive the same 2-fps sampled frames as the policy rollout. Prompts are in Appendix~\ref{app:prompts}.}
\begin{tabular}{@{}p{0.17\textwidth}p{0.18\textwidth}ccp{0.28\textwidth}@{}}
\toprule
Role & Model & Temp & max\_tok & Contract \\
\midrule
Reference caption $y^\star$ & Gemini-3.0-Pro & default & 8{,}192 & Generates the reference description for each training video \\
\midrule
Rollout claims $\mathcal{C}$ & Qwen3-VL-235B & 0.3 & 4{,}096 & Decomposes each policy caption into atomic visual claims \\
Reference claims $\mathcal{C}^{\star}$ & Gemini-3.0-Pro & default & 8{,}192 & Decomposes each reference caption into atomic visual claims \\
Video verifier & Qwen3-VL-235B & 0.3 & 4{,}096 & Labels each claim as supported, conflicting, or uncertain \\
Semantic matcher & Qwen3-VL-235B & 0.3 & 4{,}096 & Returns one-to-one semantic alignments within each category \\
\midrule
\multicolumn{5}{l}{\emph{Reward Aggregation Hyperparameters}} \\
\multicolumn{2}{l}{Precision weight $\lambda$} & \multicolumn{3}{l}{0.7 \quad (Eq.~\ref{eq:cure-reward})} \\
\multicolumn{2}{l}{Residual credit $w_u$} & \multicolumn{3}{l}{0.5 \quad (Eqs.~\ref{eq:cure-precision}--\ref{eq:cure-recall})} \\
\multicolumn{2}{l}{Over-generation guard $(\eta, \tau, \kappa)$} & \multicolumn{3}{l}{$(0.2,\; 1.5,\; 1.0)$ (Eq.~\ref{eq:overgen-regularizer})} \\
\midrule
\multicolumn{5}{l}{\emph{Reward Video Processing}} \\
\multicolumn{2}{l}{Frame sampling} & \multicolumn{3}{l}{2 fps; max 128 frames; dynamic resolution from 256$\times$256 to 768$\times$768} \\
\bottomrule
\end{tabular}
\label{tab:reward_runtime}
\end{table*}

%% file: tables/appendix/tab_category_priors.tex
\begin{table}[h]
\centering
\small
\setlength{\tabcolsep}{5pt}
\renewcommand{\arraystretch}{1.08}
\caption{Claim categories and per-category priors used by \ours.}
\begin{tabular}{p{0.35\linewidth}p{0.12\linewidth}p{0.42\linewidth}}
\toprule
Category & $\omega_k$ & Scope \\
\midrule
Action \& Interaction & 1.8 & Motion, behavior, interaction \\
Entity & 1.2 & People, animals, objects \\
Relation & 1.2 & Spatial or part-whole relations \\
Attribute & 1.0 & Color, size, shape, material, count \\
Environment & 1.0 & Scene, place, weather, background \\
Camera & 0.8 & Viewpoint, shot type, camera motion \\
Lighting & 0.8 & Light source, brightness, time of day \\
Emotion \& Atmosphere & 0.8 & Mood conveyed by visible content \\
Text & 0.8 & Visible text, signs, subtitles \\
Visibility & 0.8 & Blur, occlusion, partial visibility \\
\midrule
\multicolumn{3}{l}{\emph{Default for unrecognized categories: $1.0$}} \\
\bottomrule
\end{tabular}
\label{tab:category_priors}
\end{table}

%% file: tables/appendix/tab_direct_eval_protocol.tex
\begin{table*}[htbp]
\centering
\small
\setlength{\tabcolsep}{4pt}
\renewcommand{\arraystretch}{1.12}
\caption{Direct caption evaluation protocol for Table~\ref{tab:video_hallusion_compact_results}.}
\begin{tabular}{p{0.18\textwidth}p{0.24\textwidth}p{0.21\textwidth}p{0.29\textwidth}}
\toprule
Benchmark & What is scored & Metrics & Notes \\
\midrule
EventHallusion & Description-level caption quality on Entire, Interleave, and Misleading subsets & Entire, Interleave, Misleading, Overall & Higher is better for all scores \\
VCapsBench & Caption factuality and coverage via text-only QA & Accuracy Rate (AR), Inconsistency Rate (IR), Coverage Rate (CR) & Lower is better for IR; higher for AR and CR \\
DREAM-1K & Caption-level event precision and recall via AutoDQ extraction & Precision, Recall, F1 & Evaluates faithfulness and density jointly \\
\bottomrule
\end{tabular}
\label{tab:direct_eval_protocol}
\end{table*}

%% file: tables/appendix/tab_baseline_groups.tex
\begin{table*}[htbp]
\centering
\small
\setlength{\tabcolsep}{4pt}
\renewcommand{\arraystretch}{1.08}
\caption{Baseline grouping and result sourcing for Table~\ref{tab:video_hallusion_compact_results}. ``Reproduced'' indicates results obtained with official code and our evaluation pipeline; ``API-queried'' indicates results from proprietary model APIs.}
\begin{tabular}{p{0.22\linewidth}p{0.30\linewidth}p{0.18\linewidth}p{0.22\linewidth}}
\toprule
Group & Models & Source & Notes \\
\midrule
Proprietary references & Seed 2.0 Pro, Gemini 3.0 Pro & API-queried & Excluded from open-source bolding \\
Open-source captioners & Tarsier-34B, Tarsier2-7B, OwlCap-7B, LLaVA-Video-7B & Reproduced & Official code + default prompts \\
General-purpose VLMs & InternVL3.5-30B-A3B, Qwen2.5-VL-72B-Instruct, Qwen3-VL-30B-A3B-Instruct, Qwen3-VL-235B-A22B-Instruct, Qwen3.6-35B-A3B & Reproduced & Default captioning prompt \\
\ours & Qwen3-VL-30B-A3B-Instruct + CuRe post-training & This work & Included in open-source set \\
\bottomrule
\end{tabular}
\label{tab:baseline_groups}
\end{table*}

%% file: tables/appendix/tab_prism_protocol.tex
\begin{table*}[htbp]
\centering
\small
\setlength{\tabcolsep}{4pt}
\renewcommand{\arraystretch}{1.12}
\caption{Two-stage Prism-style caption-to-QA protocol for Table~\ref{tab:prism_model_benchmark_results}.}
\begin{tabular}{p{0.17\textwidth}p{0.23\textwidth}p{0.51\textwidth}}
\toprule
Stage & Variable & Input and output \\
\midrule
Stage 1 captioning & Captioner row in Table~\ref{tab:prism_model_benchmark_results} & The captioner receives the video and the caption prompt in Figure~\ref{fig:prompt-direct-caption}; it outputs a text caption \\
Stage 2 QA & Fixed QA model & Seed 2.0 Pro receives only the generated caption, benchmark question, and answer options, using Figure~\ref{fig:prompt-prism-qa} \\
Scoring & Fixed benchmark evaluator & The benchmark official answer key or evaluator scores the Stage~2 answer \\
\bottomrule
\end{tabular}
\label{tab:prism_protocol}
\end{table*}

%% file: tables/appendix/tab_ablation_protocol.tex
\begin{table*}[htbp]
\centering
\caption{Training and RL reward definitions for Table~\ref{tab:ablation_combined}. All SFT + GRPO rows start from the same SFT checkpoint and use the same rollout configuration; only the reward changes.}
\label{tab:ablation_protocol}
\small
\setlength{\tabcolsep}{3.5pt}
\renewcommand{\arraystretch}{1.12}
\begin{tabular}{p{0.14\textwidth}p{0.13\textwidth}p{0.25\textwidth}p{0.38\textwidth}}
\toprule
Row & Training & RL reward & Purpose \\
\midrule
SFT & SFT only & -- & Measures the warmstarted captioner before RL post-training \\
Holistic reward & SFT + GRPO & Whole-caption scalar reward without claim decomposition & Tests whether response-level RL feedback is sufficient without claim-level rubric structure \\
Base P/R & SFT + GRPO & Claim P/R, Eq.~\ref{eq:base-pr} & Tests the basic claim-level control without CuRe calibration \\
Precision-only & SFT + GRPO & Claim precision, $R=p^{\mathrm{CuRe}}$ using Eqs.~\ref{eq:aggregated-pr} and~\ref{eq:cure-precision} & Isolates the precision-side selective-credit term \\
Recall-only & SFT + GRPO & Claim recall, $R=r^{\mathrm{CuRe}}$ using Eqs.~\ref{eq:aggregated-pr} and~\ref{eq:cure-recall} & Isolates the recall-side bounded reference-target expansion term \\
Full \ours & SFT + GRPO & Claim precision + recall, Eq.~\ref{eq:cure-reward}, with Eq.~\ref{eq:overgen-regularizer} & Combined precision-recall reward used by the main model \\
\bottomrule
\end{tabular}
\end{table*}

%% file: tables/appendix/tab_downstream_protocol.tex
\begin{table*}[htbp]
\centering
\scriptsize
\setlength{\tabcolsep}{2.6pt}
\renewcommand{\arraystretch}{1.12}
\caption{Hyperparameters for the downstream pretraining utility study. Stage~2 is the only stage whose caption source changes across paired rows in Table~\ref{tab:downstream_pretraining_utility}.}
\begin{tabular}{@{}p{0.14\textwidth}p{0.18\textwidth}p{0.18\textwidth}p{0.18\textwidth}p{0.18\textwidth}@{}}
\toprule
Stage & Trainable modules & Data & Optimization & Length / video budget \\
\midrule
Initial Alignment & Projector / aligner only; ViT and LLM frozen & BLIP-3-Kale image-caption JSONL & 1 epoch; LR $1{\times}10^{-3}$; min LR $1{\times}10^{-4}$; warmup 0.05; global / micro batch 512 / 64 & Max length 8{,}192; image max token 1{,}024 \\
Further Pretraining & ViT, LLM, and projector / aligner all unfrozen & One Stage~2 caption corpus per run; original captions or \ours recaptions & 1 epoch; LR $4{\times}10^{-5}$; min LR $2{\times}10^{-6}$; warmup 0.03; global / micro batch 64 / 1 & Max length 64{,}000; video max token 1{,}024; 2 fps; at most 512 frames \\
VideoQA SFT & ViT, LLM, and projector / aligner all unfrozen & Molmo2-AskModelAnything 20K + LLaVA-Video-178K Stage-3 80K by default & 1 epoch; LR $1{\times}10^{-5}$; min LR $2{\times}10^{-6}$; warmup 0.03; global / micro batch 128 / 1 & Max length 64{,}000; same video budget as Stage~2 \\
\bottomrule
\end{tabular}
\label{tab:downstream_protocol}
\end{table*}

%% file: prompt_templates/reward_decompose_en_box.tex
\begingroup
\scriptsize\ttfamily\raggedright
\noindent You are a visual information analysis assistant. Decompose the following video/image description into independent atomic visual facts.\par
\medskip\par
\noindent \#\#\# Rules\par
\noindent 1. **Atomicity**: Each fact describes exactly one directly observable property or phenomenon. If a sentence covers an entity, its attribute, and an action, split them into separate facts.\par
\noindent 2. **Explicit subjects**: Use original nouns (e.g., "the cat", "the red car"). Never use pronouns like "it" or "this". Use the full description on first mention, then a shorter form.\par
\noindent 3. **No redundancy**: Do not repeat the same attribute across facts.\par
\noindent 4. **Temporal \& dynamic**: Camera movements (pan, tilt, zoom), scene transitions, and motion trajectories should each be an independent fact.\par
\noindent 5. **Category constraint**: The category field must be one of the following enum values:\par
\noindent    [Entity, Attribute, Relation, Action \& Interaction, Environment, Lighting, Camera, Emotion \& Atmosphere, Text, Visibility]\par
\medskip\par
\noindent \#\#\# Category Reference\par
\noindent $|$ Enum Value $|$ Scope $|$\par
\noindent $|$------------$|$-------$|$\par
\noindent $|$ Entity $|$ People, animals, objects appearing in the scene $|$\par
\noindent $|$ Attribute $|$ Color, size, shape, material, quantity of entities $|$\par
\noindent $|$ Relation $|$ Spatial relations (left of, in front of), part-of relations $|$\par
\noindent $|$ Action \& Interaction $|$ Movements, behaviors, interactions between entities $|$\par
\noindent $|$ Environment $|$ Scene, location, weather, background elements $|$\par
\noindent $|$ Lighting $|$ Light sources, direction, brightness, time of day $|$\par
\noindent $|$ Camera $|$ Camera movement, angle changes, filming techniques $|$\par
\noindent $|$ Emotion \& Atmosphere $|$ Mood, atmosphere conveyed by the visuals $|$\par
\noindent $|$ Text $|$ Visible text, signs, subtitles in the frame $|$\par
\noindent $|$ Visibility $|$ Occlusion, blur, image quality, partial visibility $|$\par
\medskip\par
\noindent \#\#\# Example\par
\noindent Description: An aerial video of a city at night. The camera slowly pushes forward, overlooking the entire city. Tall buildings fill the frame with colorful lights. Traffic flows on the roads, headlights forming light trails. A Ferris wheel with blue lights stands on the right side. The overall tone is warm, conveying a bustling urban atmosphere.\par
\medskip\par
\noindent Decomposition:\par
\noindent \textasciigrave{}\textasciigrave{}\textasciigrave{}json\par
\noindent [\par
\noindent     \{"category": "Environment", "fact": "The scene shows a city at night", "is\_visual": "yes"\},\par
\noindent     \{"category": "Camera", "fact": "The camera slowly pushes forward with an overhead perspective", "is\_visual": "yes"\},\par
\noindent     \{"category": "Entity", "fact": "There are multiple tall buildings in the frame", "is\_visual": "yes"\},\par
\noindent     \{"category": "Attribute", "fact": "The buildings are lit with colorful lights", "is\_visual": "yes"\},\par
\noindent     \{"category": "Entity", "fact": "There is traffic on the roads", "is\_visual": "yes"\},\par
\noindent     \{"category": "Action \& Interaction", "fact": "Vehicles are moving along the roads", "is\_visual": "yes"\},\par
\noindent     \{"category": "Attribute", "fact": "The headlights form continuous light trails", "is\_visual": "yes"\},\par
\noindent     \{"category": "Relation", "fact": "A Ferris wheel is positioned on the right side of the frame", "is\_visual": "yes"\},\par
\noindent     \{"category": "Lighting", "fact": "The Ferris wheel is lit with blue lights", "is\_visual": "yes"\},\par
\noindent     \{"category": "Lighting", "fact": "The overall color tone of the scene is warm", "is\_visual": "yes"\},\par
\noindent     \{"category": "Emotion \& Atmosphere", "fact": "The scene conveys a bustling urban atmosphere", "is\_visual": "yes"\}\par
\noindent ]\par
\noindent \textasciigrave{}\textasciigrave{}\textasciigrave{}\par
\medskip\par
\noindent \#\#\# Description to Decompose\par
\noindent \{caption\}\par
\medskip\par
\noindent Briefly analyze the description, then output the decomposition in the following JSON format:\par
\noindent \textasciigrave{}\textasciigrave{}\textasciigrave{}json\par
\noindent [\par
\noindent   \{"category": "$<$enum value$>$", "fact": "$<$complete independent visual fact$>$", "is\_visual": "$<$yes or no$>$"\},\par
\noindent   ...\par
\noindent ]\par
\noindent \textasciigrave{}\textasciigrave{}\textasciigrave{}\par
\noindent Ensure valid JSON format. All field values must be strings. Do not include comments in JSON.\par
\endgroup

%% file: prompt_templates/reward_verify_en_box.tex
\begingroup
\scriptsize\ttfamily\raggedright
\noindent Watch the video/image carefully and judge whether each of the following visual facts is accurate.\par
\medskip\par
\noindent \#\#\# Judgment Criteria\par
\noindent - **support**: The fact can be directly observed in the video/image. For videos, appearing at any point in time counts as support.\par
\noindent - **conflict**: The fact explicitly contradicts the visible content (e.g., wrong color, entity absent, opposite direction).\par
\noindent - **uncertain**: Cannot be confirmed or denied due to occlusion, blur, out-of-frame content, or insufficient information.\par
\medskip\par
\noindent \#\#\# Judgment Principles\par
\noindent 1. **Independence**: Judge each fact independently. Do not let the judgment of one fact influence another.\par
\noindent 2. **Visual evidence only**: Base judgments solely on directly observable content. Do not use logical reasoning or common-sense inference.\par
\noindent 3. **Conservative support**: Only mark as support when the visual content clearly confirms the fact. When in doubt, use uncertain.\par
\medskip\par
\noindent \#\#\# Visual Facts to Verify\par
\noindent \textasciigrave{}\textasciigrave{}\textasciigrave{}json\par
\noindent \{visual\_facts\}\par
\noindent \textasciigrave{}\textasciigrave{}\textasciigrave{}\par
\medskip\par
\noindent Reply in JSON format:\par
\noindent \textasciigrave{}\textasciigrave{}\textasciigrave{}json\par
\noindent [\par
\noindent     \{\par
\noindent       "category": "$<$fact category$>$",\par
\noindent       "fact": "$<$complete visual fact$>$",\par
\noindent       "analysis": "$<$specific analysis based on video/image content$>$",\par
\noindent       "result": "$<$support or conflict or uncertain$>$"\par
\noindent     \},\par
\noindent     ...\par
\noindent ]\par
\noindent \textasciigrave{}\textasciigrave{}\textasciigrave{}\par
\noindent Ensure valid JSON format. All field values must be strings. Do not include comments in JSON.\par
\endgroup

%% file: prompt_templates/reward_category_match_en_one_to_one_box.tex
\begingroup
\scriptsize\ttfamily\raggedright
\noindent You are a semantic matching expert. For the **\{category\}** category below, determine whether each "predicted claim" semantically corresponds to a "reference claim" describing the same visual fact.\par
\medskip\par
\noindent \#\#\# Matching Rules\par
\noindent 1. **Semantic equivalence**: Two claims "match" if they refer to the same specific visual fact, even with different wording or detail level.\par
\noindent 2. **Directional constraint**: Each predicted claim matches at most one reference claim; however, the same reference claim may be matched by at most one predicted claim.\par
\noindent 3. **No match**: If no semantically equivalent reference claim exists, set gt\_index to null and confidence to 0.0.\par
\noindent 4. **Reason first**: The reasoning field must be filled before confidence --- use a concise phrase ($\leq$15 words) extracting subject + action + difference.\par
\noindent 5. **Complete output**: The output array must contain exactly \{num\_pred\} elements, one per predicted claim.\par
\medskip\par
\noindent \#\#\# Confidence Scoring Guide\par
\noindent $|$ Score Range $|$ Meaning $|$ Example $|$\par
\noindent $|$-------------$|$---------$|$---------$|$\par
\noindent $|$ 0.9-1.0 $|$ Same fact, different wording $|$ "vehicles moving on the road" $\leftrightarrow$ "cars driving on the street" $|$\par
\noindent $|$ 0.7-0.8 $|$ Same core fact, detail differences do not affect fact identification $|$ "a large brown dog" $\leftrightarrow$ "a brown Labrador" $|$\par
\noindent $|$ 0.5-0.6 $|$ Same core entity/action, but missing key distinguishing attributes $|$ "a cat" $\leftrightarrow$ "a black cat with white paws" $|$\par
\noindent $|$ 0.0-0.4 $|$ Different facts or unrelated $|$ "the sky is blue" $\leftrightarrow$ "a dog is running" $|$\par
\medskip\par
\noindent **Important rule**: If the predicted claim omits key distinguishing attributes (color, quantity, size, material, etc.) from the reference claim, confidence should NOT exceed 0.5.\par
\medskip\par
\noindent ---\par
\medskip\par
\noindent \#\#\# Predicted Claims (\{num\_pred\} items)\par
\noindent \{pred\_claims\_json\}\par
\medskip\par
\noindent \#\#\# Reference Claims (\{num\_gt\} items)\par
\noindent \{gt\_claims\_json\}\par
\medskip\par
\noindent \#\#\# Output Requirement\par
\noindent Output a JSON array containing exactly **\{num\_pred\}** elements, one per predicted claim:\par
\noindent \textasciigrave{}\textasciigrave{}\textasciigrave{}json\par
\noindent [\par
\noindent   \{"pred\_index": 0, "gt\_index": "$<$int or null$>$", "reasoning": "$<$$\leq$15 word concise reasoning$>$", "confidence": "$<$0.0-1.0$>$"\},\par
\noindent   ...\par
\noindent ]\par
\noindent \textasciigrave{}\textasciigrave{}\textasciigrave{}\par
\endgroup